\documentclass{article}
\usepackage[preprint]{neurips_2026}

\usepackage[utf8]{inputenc}
\usepackage[T1]{fontenc}
\usepackage{url}
\usepackage{booktabs}
\usepackage{amsmath,amssymb,amsfonts}
\usepackage{xcolor}
\usepackage{graphicx}
\usepackage{float}
\usepackage{hyperref}
\graphicspath{{figs/}}

\newcommand{\rhov}{\rho}

\title{Does Depth Survive Composition, and Can a Router Use It?\\
Adaptive Compute in Latent World Models}

\author{%
  Achyuthan Sivasankar\\
  New York University\\
  \texttt{as21154@nyu.edu}
}

\begin{document}
\maketitle

\begin{abstract}
Adaptive-compute world models run fewer predictor blocks on some rollout steps than on others. The
design rests on two assumptions: that a deeper predictor gives a more accurate rollout, and that the
steps which need depth can be identified before the predictor runs. We test both on nine DeepMind
Control tasks with a twelve-block early-exit latent predictor, eight seeds per task. The first
assumption fails on three tasks: there, a predictor cut at a shallower exit rolls out more accurately
than the full stack on every seed. Composing predictions over ten steps also moves the depth--error
relation away from its one-step shape, in both directions. In hindsight, choosing depth per step
helps: a greedy oracle that sees the true next latent has 10--37\% lower rollout error than the best
mixture of fixed depths at the same average compute, and 8--28\% lower than the best step-index
schedule, on all nine tasks. No router we trained captures this. The difficulty router these models
carry stays within 5\% of the fixed-depth mixtures. A router trained to predict each exit's error
beats them by up to 12\%, but no more than a fixed schedule does, and picks the best exit no more
often than always guessing the most common one. In closed-loop planning at equal compute, the routed
planner matches a fixed middle depth. On cheetah, where depth 2 is the worse one-step predictor but the better rollout operator,
supervising the early exits only at the first rollout step leaves depth 2's one-step error
unchanged, raises its rollout error by 16\% and removes the inversion. A control with the same loss
weight that supervises the exits at one random rollout step keeps most of the inversion ($\rhov$
$0.91$ against $0.85$ standard and $1.14$ first-step-only, three seeds), so the inversion depends on
supervising the exits from composed latents. These results change with the encoder,
backbone, rollout horizon and training data, which we map. The fixed-depth campaign and its decision
rules were pre-registered; the routing analyses, the equal-compute planner runs, the ablation
control and the one-step decomposition were added later and are exploratory.
\end{abstract}

\section{Introduction}
A latent world model encodes an observation to a latent $z$ and predicts future latents
$z_{t+1}=f(z_t,a_t)$. Planning and model-based control then roll $f$ forward autoregressively,
calling it tens to hundreds of times per encoded observation
\citep{ha2018world,hafner2019planet,hansen2024tdmpc2}. A model that closes a control loop at tens of
hertz therefore spends most of its compute in the predictor, not the encoder. Running fewer predictor
blocks on some steps is a natural way to make planning cheaper, and it brings early-exit and
mixture-of-depths machinery from supervised and language models into world models
\citep{teerapittayanon2016branchynet,graves2016act,raposo2024mod,lu2026loopedwm}.

That design rests on two assumptions, and they are separate. The first is that depth helps under
rollout: a deeper predictor should give a more accurate \emph{rollout}, not only a more accurate
single step. The second is that the need for depth can be anticipated: a router has to tell, from the
current latent and action, which steps deserve more blocks. The first is a statement about fixed
depths. The second is about choosing depth per step, and it can hold or fail whatever the answer to
the first. If shallow rollouts are better on average, a router can still gain by sending the steps
that need depth to the deep exit. If deep rollouts are better on average, a router gains only if it
can find the steps where they are not.

We measure each assumption with its own instrument, on nine DeepMind Control (DMC) tasks
\citep{tassa2018dmc,todorov2012mujoco} with a twelve-block latent predictor trained under matched
single-step and four-step objectives, eight seeds each. For fixed depths we report the \emph{shallow
penalty} $\rhov=\mathrm{err}(\text{shallowest-exit rollout})/\mathrm{err}(\text{full-depth
rollout})$ (Section~\ref{sec:setup}) together with the full per-depth curve, at one step and after a
ten-step rollout. For routing we compare, at equal average blocks per step, three routers, random
routing and a per-step oracle against mixtures of fixed depths and against step-index schedules
(Section~\ref{sec:routing}), and repeat the comparison inside a planner (Section~\ref{sec:planner}).

\begin{figure}[t]
\centering
\includegraphics[width=\linewidth]{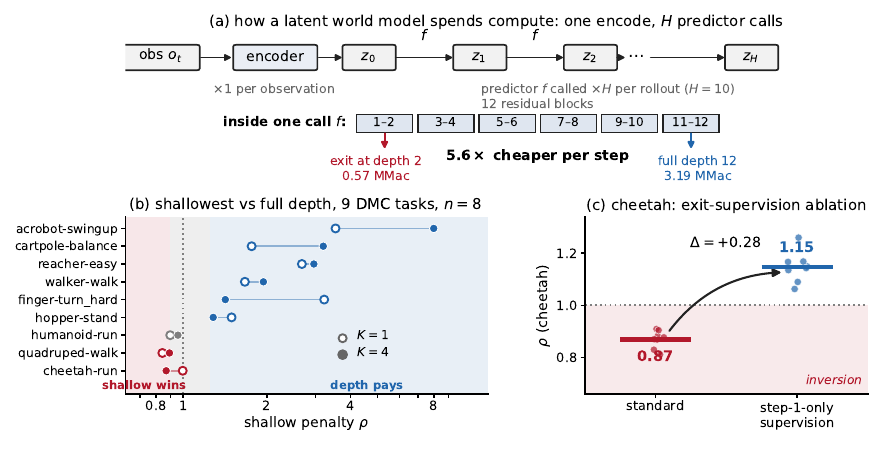}
\caption{\textbf{Setup and the two fixed-depth results.}
\textbf{(a)} The encoder runs once per observation; the predictor $f$ runs once per rollout step,
so predictor depth sets the cost of planning. Exiting at depth 2 instead of 12 is $5.6\times$
cheaper per step.
\textbf{(b)} Shallowest-exit over full-depth rollout error for nine tasks ($K{=}1$ open, $K{=}4$
filled, eight seeds). Below 1, the depth-2 rollout is more accurate than the depth-12 one. This
compares two fixed depths only; the full depth curves are in Figure~\ref{fig:onestep}, and
Section~\ref{sec:routing} asks whether choosing depth per step helps.
\textbf{(c)} On cheetah, supervising the exits only at the first rollout step raises the ratio from
$0.87$ to $1.15$ ($n{=}8$, non-overlapping); supervising them at one random step instead keeps most
of the inversion (Section~\ref{sec:catch22}).}
\label{fig:overview}
\end{figure}

\paragraph{Contributions.} (i) Per-depth rollout and one-step error for nine tasks, two training
objectives and eight seeds, with absolute errors and persistence baselines
(Section~\ref{sec:taxonomy}). (ii) An equal-compute comparison of three routers, random
routing and a greedy oracle against fixed-depth mixtures and searched step-index schedules, in open
loop (Section~\ref{sec:routing}) and in a planner against every fixed exit
(Section~\ref{sec:planner}). (iii) An exit-supervision ablation on cheetah, decomposed into
one-step and rollout error at both ends of the network (Section~\ref{sec:catch22}). (iv) How the
encoder, backbone, horizon, metric and training data change these results
(Section~\ref{sec:cautions}), and whether the fixed-depth ordering carries into closed-loop planning (Section~\ref{sec:planner}).

\section{Related work}
\paragraph{Adaptive computation.} Early-exit networks \citep{teerapittayanon2016branchynet,
figurnov2017spatially,bolukbasi2017adaptive}, Adaptive Computation Time and its successors
\citep{graves2016act,dehghani2019universal,banino2021pondernet}, input-adaptive depth in transformers
\citep{elbayad2020depth,raposo2024mod} and mixture-of-experts routing
\citep{shazeer2017moe,fedus2022switch} allocate compute per input \citep{han2021dynamic}. Two results
are closest to ours. \citet{kaya2019shallowdeep} show that in classification an internal exit is
often correct when the final layer is wrong, which they call overthinking; we find the same effect in
one-step latent prediction on some tasks (Section~\ref{sec:taxonomy}). \citet{schuster2022calm}
study early exit during autoregressive text generation, where an exit taken at one token changes the
inputs to every later token, and calibrate per-token exits against a sequence-level quality
constraint. That is the structural problem studied here: per-step exits composed over a sequence.
Our object is different (a latent dynamics predictor rolled out by a planner, not a decoder), and so
is our question. CALM asks how to exit safely given that depth helps; we measure whether depth helps
after composition, and whether per-step choice helps at equal compute.

\paragraph{Latent world models and planning.} World models learn a dynamics function and plan or
imagine within it \citep{ha2018world,schrittwieser2020muzero,chua2018pets,janner2019mbpo}. JEPA-style
latent models \citep{lecun2022path,assran2023ijepa,bardes2024vjepa} and latent-space planners
\citep{hafner2020dream,hafner2023dreamerv3,hansen2024tdmpc2} roll a learned latent dynamics function
forward under candidate action sequences. \citet{lu2026loopedwm} build a world model around a
parameter-shared block applied a variable number of times and report computation that adapts depth to
each prediction step. Systems of that kind depend on two things we measure directly on a small,
controlled testbed: that more depth yields a better rollout, and that the steps needing depth can be
identified in advance.

\paragraph{Compounding error and evaluation.} One-step model errors compound over a rollout, and a
long line of work studies that growth and trains against it
\citep{talvitie2014hallucinated,venkatraman2015dad,asadi2018lipschitz,lambert2022compounding};
latent-overshooting and multi-step consistency losses
\citep{hafner2019planet,hafner2020dream,hansen2022tdmpc,schwarzer2021spr,gelada2019deepmdp} are the
form this takes in latent world models. We use a controlled $K{=}1$ versus $K{=}4$ overshoot as an
experimental knob and ask how compounding interacts with predictor depth. Latent models are usually
scored by prediction error against a stop-gradient target \citep{grill2020byol,chen2021simsiam}; on
one task our conclusion changes between latent-space and observation-space error, in line with
broader calls for evaluation care in deep RL \citep{henderson2018matters,agarwal2021rliable}.

\section{Setup and instrument}\label{sec:setup}
\paragraph{Model.} We use a block-stacked latent predictor with early-exit heads at depths
$\{2,4,\dots,12\}$ on top of an MLP encoder with an EMA target \citep{grill2020byol}, trained with a
latent JEPA loss \citep{assran2023ijepa}, deep supervision of the shallow exits
\citep{lee2015deeply,szegedy2015going}, light distillation toward the full-depth output
\citep{hinton2015distilling}, and VICReg regularization \citep{bardes2022vicreg}
(Appendix~\ref{app:hp}). This is a standard adaptive-compute world-model backbone; our claims concern
its rollout behavior, not the architecture.

\paragraph{Why state-based DMC.} We study proprioceptive DMC rather than pixels for the main campaign
by design: our own results (Section~\ref{sec:cautions}) show the \emph{encoder} moves the regime, so
a pixel pipeline would confound the depth--composition question with representation-learning quality.
We therefore fix the simplest competent encoder for the controlled study, then \emph{test} pixels as
one of five perturbation axes.

\paragraph{The shallow penalty.} For a checkpoint, we encode held-out windows to $z_0$, roll the
predictor $H{=}10$ steps at each \emph{fixed} exit depth $d$, and measure the horizon-mean latent
error $e_d$ against the EMA-encoded true future. The full-depth error $e_{\mathrm{full}}$ is the
denominator and
\[
\rhov \;=\; \frac{e_{d_{\min}}}{e_{\mathrm{full}}},\qquad d_{\min}=\text{shallowest exit},
\]
with per-depth curves $e_d/e_{\mathrm{full}}$ retained for shape. $\rhov$ is computed by a single
audit script over 256 held-out windows (seed fixed); a collapse guard rejects any checkpoint whose
$e_{\mathrm{full}}$ is numerically degenerate (none were).

\paragraph{Compute accounting and the multi-step knob.} Each residual predictor block
($d_{\text{model}}{=}256$, MLP ratio $2$) costs $\approx0.26$\,MMac, so the shallowest exit (depth 2,
$0.57$\,MMac) is $\mathbf{5.6\times}$ cheaper than the full stack (depth 12, $3.19$\,MMac). Because the predictor is called once per \emph{rollout step} while the encoder runs once per
observation, depth routing scales almost linearly with planning FLOPs; $\rhov$ is the factor by which
spending $5.6\times$ fewer blocks per step changes rollout error. We train
matched twins differing only in the training horizon $K$: $K{=}1$ (single-step) and $K{=}4$ (latent
overshooting, back-propagating through a four-step full-depth rollout against EMA targets). Both
twins draw identical, model-independent, pre-collected data in identical minibatch order, so a
$K{=}1$/$K{=}4$ pair differs only in the loss.

\paragraph{The router.} Every model carries a difficulty head: a two-layer MLP with 128 hidden units
that maps $(z_t,a_t)$ to $d_t\in[0,1]$. It is trained on stop-gradient inputs to regress the model's
own full-depth one-step error, after a running standardisation and a sigmoid. A parameter-free
controller turns $d_t$ into a depth in one of two ways. The \emph{graded} rule splits $[0,1]$ evenly
across the six exits. The \emph{binary} rule runs depth 12 when $d_t\ge\tau$ and depth 2 otherwise.
The planning experiments of Section~\ref{sec:planner} use the binary rule with $\tau{=}0.5$. Note what
the head is trained on: how large the full-depth error will be, not which exit will roll out best.

\paragraph{Routing at equal compute.} $\rhov$ compares two fixed depths. To ask whether choosing
depth per step helps, we compare policies at equal average blocks per step on the same held-out
windows. Running a fraction of rollouts at one fixed depth and the rest at another costs the
weighted average of the two depths and has the weighted average of their errors, so every point on
the lower convex hull of the six fixed-depth points is achievable. We call that hull the
\emph{fixed-depth frontier} and score every policy by its error over the frontier's error at no more
than the policy's average blocks per step. The frontier is not the best policy that ignores the
state: a \emph{schedule}, one depth per step index applied to every window, can do better, because
it can spend depth on some steps of every rollout instead of on all steps of some rollouts. We
therefore also search schedules: every two-phase schedule (depth $a$ for the first $k$ steps, depth
$b$ after) and a greedy path that starts at depth 2 and repeatedly deepens the step with the largest
error drop per added block. Schedules are chosen on 256 windows from separate validation episodes and
scored on the evaluation windows; the lower hull of the chosen schedules is the \emph{schedule
frontier}. A policy uses the state, rather than the step index, only if it beats the schedule
frontier.

Against these references we place five policies. (i) The trained difficulty router, under both rules.
(ii) \emph{Random routing}, which picks depth 2 or 12 independently at each step with a fixed
probability: the same two choices the binary rule has. (iii) A greedy per-step \emph{oracle} that sees
the true next latent and takes the lowest-error exit at each step, together with a family that takes
the cheapest exit within a tolerance of that step's full-depth error. The oracle is greedy, so it
bounds what a router with one-step lookahead could do, not the best possible depth sequence. (iv) An
\emph{exit-error router} trained for the routing decision itself: an MLP that maps $(z_t, a_t, t/H)$
to each exit's log error relative to full depth and takes the exit minimising predicted error plus
$\lambda\cdot$depth. It is trained on states visited by full-depth, oracle and random-exit rollouts of
4096 windows from 30 separate episodes, plus one round on its own rollouts, with early stopping on
validation episodes. We sweep $\lambda$
and choose it per held-out seed from the task's other seeds. (v) An \emph{end-to-end router} trained
through the rollout with a penalty that holds its mean depth at the graded router's budget, in the
style of \citet{graves2016act} and \citet{raposo2024mod}, hard-routed at evaluation.

\paragraph{Pre-registration.} Before running the compute campaign we fixed and committed the regime
thresholds, two decision gates (A, B), a metric-artifact kill criterion, and the precedence rules
among them; each is read once, mechanically, from persisted JSON. Writing $\rhov^{K}$ for the seed
mean at training horizon $K$, a task is \textbf{intrinsic} if $\rhov^{K1}\ge1.25$; otherwise, when
$\rhov^{K1}<1.15$, it is \textbf{inversion} if $\rhov^{K4}<0.90$, \textbf{flat} if
$0.90\le\rhov^{K4}<1.25$, and \textbf{created} if $\rhov^{K4}\ge1.25$; anything else is
\textbf{boundary}. (The \emph{created} regime---a tradeoff manufactured by multi-step training---was
the outcome the study was originally designed to find.) Several analyses in this paper were
\emph{not} pre-registered: the equal-compute routing comparison, schedule search and trained routers
of Section~\ref{sec:routing}, the one-step decomposition and the supervision-matched control of
Section~\ref{sec:catch22}, and the equal-compute planner runs of Section~\ref{sec:planner}. They were
added after the reviews of an earlier version and should be read as exploratory. All except the
control, which trains new models, use the frozen checkpoints of the pre-registered campaign.
Appendix~\ref{app:repro} describes how each number is produced.

\section{The hypothesis that failed}\label{sec:negative}
The study began as a directed test of a specific mechanism. Unpublished pilot runs of ours,
at reduced scale, had shown that
single-step-trained routers save essentially no compute in rollouts, because shallow and deep
rollouts were near-indistinguishable ($\rhov\approx1$)---routing was vacuous. We hypothesized this is
an artifact of the single-step loss and that multi-step training would \emph{create} a depth--quality
tradeoff, pre-registering it as \textbf{Gate 1} (pass iff $\ge 2/3$ of the three tasks first
available are \emph{created}). \textbf{Gate 1 failed, 0/3}: cheetah \emph{inverted}
($\rhov^{K4}=0.85$ at three seeds), humanoid stayed flat, and walker already had a large tradeoff at $K{=}1$
(intrinsic, not created). A reduced-scale pilot showing an encouraging $1.25\times$ on cheetah
collapsed to $0.85\times$ at full scale---the reduced-scale mirage the full-scale pre-registration
was designed to catch. Rather than discard the machinery, we broadened to nine tasks and asked the
descriptive question this paper answers: \emph{what regimes actually occur, and why?}

\section{A taxonomy of depth-composition regimes}\label{sec:taxonomy}
We trained the six additional DMC tasks under the same matched $K{=}1/K{=}4$ design and later
expanded every cell to \textbf{eight seeds} ($9\times2\times8=144$ runs; all healthy---every checkpoint beats a
persistence baseline, none collapsed) and classified all nine (Figure~\ref{fig:overview}b,
Table~\ref{tab:taxonomy}).

\begin{table}[tb]
\centering
\small
\begin{tabular}{lccl}
\toprule
task & $\rhov^{K1}$ [95\% CI] & $\rhov^{K4}$ [95\% CI] & regime\\
\midrule
acrobot-swingup    & 3.54 [2.96, 4.21] & 7.98 [6.78, 9.54] & intrinsic\\
finger-turn\_hard  & 3.22 [2.16, 4.33] & 1.42 [1.23, 1.65] & intrinsic\\
reacher-easy       & 2.68 [2.20, 3.23] & 2.95 [2.51, 3.34] & intrinsic\\
cartpole-balance   & 1.76 [1.55, 2.03] & 3.19 [2.65, 3.75] & intrinsic\\
walker-walk        & 1.67 [1.61, 1.72] & 1.95 [1.89, 2.00] & intrinsic\\
hopper-stand       & 1.49 [1.42, 1.59] & 1.28 [1.27, 1.29] & intrinsic\\
humanoid-run       & 0.90 [0.89, 0.91] & 0.96 [0.95, 0.97] & flat\\
cheetah-run        & 1.00 [0.96, 1.03] & 0.87 [0.85, 0.89] & inversion\\
quadruped-walk     & 0.84 [0.83, 0.85] & 0.89 [0.88, 0.90] & inversion\\
\midrule
\multicolumn{4}{l}{\textbf{Gate A: PARTIAL} --- intrinsic 6, inversion 2, flat 1, boundary 0.}\\
\bottomrule
\end{tabular}
\caption{\textbf{Nine-task taxonomy} (eight-seed mean $\rhov$ with bootstrap 95\% CIs, $H{=}10$, 256
held-out windows). Two of the three core regimes hold $\ge2$ tasks, so the pre-registered
\textbf{Gate A} (\textsc{pass} iff all three core regimes hold $\ge2$ tasks, \textsc{partial} iff
two do) returns \textsc{partial}: an intrinsic-vs-inversion dichotomy with a flat singleton.
\textbf{Expanding from three to eight seeds changed no task's label.} No task is boundary.}
\label{tab:taxonomy}
\end{table}

\paragraph{The intrinsic tradeoff is the majority regime.} On 6/9 tasks $\rhov^{K1}\ge1.25$: depth
buys rollout quality \emph{already under single-step training}. This recontextualizes the failed
hypothesis of Section~\ref{sec:negative}: ``latent rollouts don't reward depth'' was a
\emph{task-selection artifact} of the two tasks first studied, cheetah and humanoid being precisely
the two non-intrinsic outliers. Multi-step training's effect on intrinsic tasks is
mixed---it amplifies some (acrobot $3.54\to7.98$, reacher, cartpole) and \emph{reduces} others
(finger $3.22\to1.42$, hopper $1.49\to1.28$)---so ``train longer horizons'' is not a reliable lever.

\paragraph{Seed stability, curve shapes, and borderline cases.} At eight seeds the task labels are
stable: expanding from three seeds changed \emph{no} label, and CIs are tight. Individual seeds are
less settled near the $0.90$ threshold. Per-seed labels agree with the task label on 7/9 tasks; on
cheetah and quadruped two of eight seeds each sit just above $0.90$ and read as flat, and quadruped's
CI reaches the threshold (upper bound $0.901$). Every seed of both tasks is non-intrinsic. The per-depth curves (Appendix~\ref{app:grid}) show the
three shapes: intrinsic error falls with depth; inversion curves dip \emph{below} 1 at intermediate
depths; flat curves hug 1. Humanoid, flat by $\rhov$ (the ratio at the \emph{shallowest} exit), dips
to $\approx0.84$ at intermediate depth---inversion-shaped, and its label flips under the
metric-space control. Cheetah is the most
\emph{robust} inversion (six of eight seeds below the $0.90$ threshold, the other two at $0.905$ and
$0.909$; both metric spaces; dipping to $0.77$ at mid-depth); we use it as the primary object of the
mechanistic analysis below.

\begin{figure}[tb]
\centering
\includegraphics[width=0.86\linewidth]{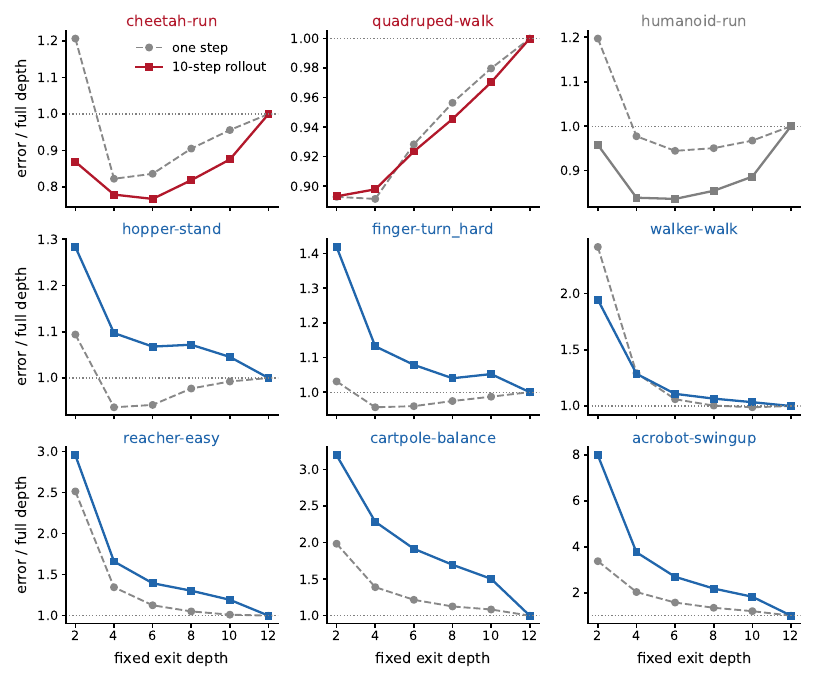}
\caption{\textbf{Error at every exit relative to full depth}, at one step (dashed) and averaged over a
ten-step rollout (solid), $K{=}4$, eight-seed means, ordered by $\rhov$. Rollout moves the curve away
from its one-step shape in both directions.}
\label{fig:onestep}
\end{figure}

\paragraph{The full depth curve, at one step and after rollout.} $\rhov$ looks only at the shallowest
exit, and the reviewers of an earlier version were right that this hides most of the curve.
Table~\ref{tab:depth} and Figure~\ref{fig:onestep} report every exit, at one step (each step
starting from the encoded true state) and after a ten-step rollout. Three tasks have a best fixed
depth below 12 on all eight seeds: cheetah (depth 6, $0.77\times$ the full-depth rollout error),
humanoid (depth 6, $0.84\times$) and quadruped (depth 2, $0.89\times$). Humanoid is labelled flat
by $\rhov$ but has a clear minimum at depth 6. Some middle exit beats the full stack at one step on six
tasks, by 4--18\% on five of them and by 1\% on walker, in line with the overthinking effect
\citet{kaya2019shallowdeep} report for classifiers.
Composition then moves the curve in both directions. On hopper and finger the middle exits win at
one step and lose after rollout. On cheetah depth 2 is the worst exit at one step
($1.21\times$) and better than full depth after rollout ($0.87\times$). On reacher, cartpole and
acrobot rollout widens a gap that already exists at one step; on walker it narrows the depth-2 gap
($2.42\to1.95$). All models beat persistence
by a wide margin: full-depth rollout error is $76$--$99.8\%$ below that of holding the current latent
fixed (Appendix~\ref{app:abs}, Table~\ref{tab:abs}).

\begin{table}[tb]
\centering
\small
\setlength{\tabcolsep}{4pt}
\resizebox{\linewidth}{!}{%
\begin{tabular}{llccccc}
\toprule
task & regime & $\rhov_1$ (one step) & $\rhov$ (rollout) & best depth & its ratio & seeds best${<}12$\\
\midrule
acrobot-swingup & intrinsic & $3.37$ $[3.00, 3.78]$ & $7.98$ $[6.78, 9.54]$ & 12 & $1.00$ $[1.00, 1.00]$ & 0/8\\
cartpole-balance & intrinsic & $1.98$ $[1.88, 2.16]$ & $3.19$ $[2.65, 3.75]$ & 12 & $1.00$ $[1.00, 1.00]$ & 1/8\\
reacher-easy & intrinsic & $2.51$ $[2.16, 2.87]$ & $2.95$ $[2.51, 3.34]$ & 12 & $1.00$ $[1.00, 1.00]$ & 1/8\\
walker-walk & intrinsic & $2.42$ $[2.38, 2.45]$ & $1.95$ $[1.89, 2.00]$ & 12 & $1.00$ $[1.00, 1.00]$ & 0/8\\
finger-turn\_hard & intrinsic & $1.03$ $[0.98, 1.08]$ & $1.42$ $[1.23, 1.65]$ & 12 & $1.00$ $[1.00, 1.00]$ & 5/8\\
hopper-stand & intrinsic & $1.09$ $[1.06, 1.14]$ & $1.28$ $[1.27, 1.29]$ & 12 & $1.00$ $[1.00, 1.00]$ & 1/8\\
humanoid-run & flat & $1.20$ $[1.19, 1.21]$ & $0.96$ $[0.95, 0.97]$ & 6 & $0.84$ $[0.83, 0.84]$ & 8/8\\
quadruped-walk & inversion & $0.89$ $[0.88, 0.90]$ & $0.89$ $[0.88, 0.90]$ & 2 & $0.89$ $[0.88, 0.90]$ & 8/8\\
cheetah-run & inversion & $1.21$ $[1.17, 1.25]$ & $0.87$ $[0.85, 0.89]$ & 6 & $0.77$ $[0.74, 0.80]$ & 8/8\\
\bottomrule
\end{tabular}}
\caption{\textbf{Depth at one step and after rollout} ($K{=}4$, eight-seed means, bootstrap 95\%
CIs). $\rhov_1$ and $\rhov$ are depth-2 over depth-12 error at one step and averaged over a ten-step
rollout. ``Best depth'' is the exit with the lowest mean rollout error and ``its ratio'' that error
over full depth. The last column counts seeds whose own best exit is shallower than 12.}
\label{tab:depth}
\end{table}

\section{Choosing depth per step at equal compute}\label{sec:routing}
The comparisons so far use fixed depths. We now ask whether choosing depth per step helps, using the
policies defined in Section~\ref{sec:setup}, on the same 256 held-out windows and $K{=}4$ models.
(The audit recomputes the fixed-depth errors from the checkpoints and reproduces every stored
$\rhov$ in Table~\ref{tab:taxonomy}, all 158 checkpoints, exactly.) Throughout, a policy is scored
by its rollout error divided by the best error a mixture of fixed depths can reach with at most the
same average number of blocks. Below 1, the policy does better than any mixture of
fixed depths.

\begin{figure}[tb]
\centering
\includegraphics[width=\linewidth]{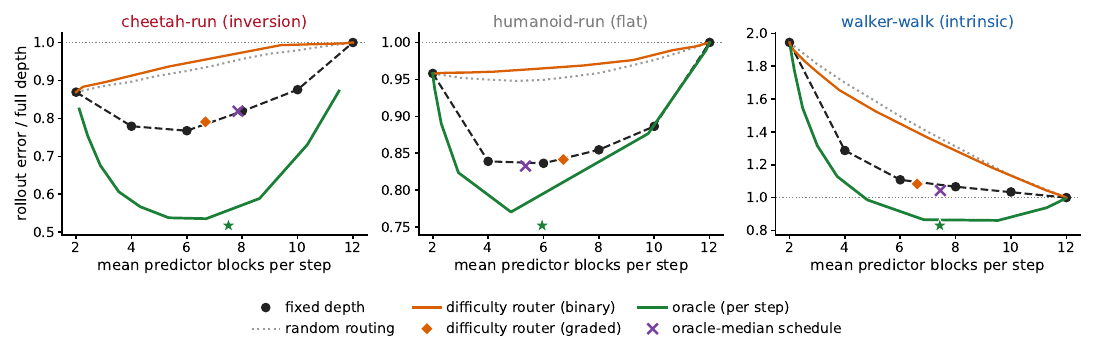}
\caption{\textbf{Rollout error against compute}, $K{=}4$, eight-seed means, relative to full depth.
Dashed: mixtures of fixed depths (the fixed-depth frontier). Green: the greedy oracle family,
with the star at its lowest-error member. Orange: the difficulty router, binary rule swept over $\tau$
and the graded rule (diamond). Purple cross: the oracle's median depth per step, applied as a schedule. Dotted grey: random routing between depths 2 and 12. All nine tasks
are in Appendix~\ref{app:routing}.}
\label{fig:frontier}
\end{figure}

\begin{table}[tb]
\centering
\small
\setlength{\tabcolsep}{3pt}
\resizebox{\linewidth}{!}{%
\begin{tabular}{lccccccc}
\toprule
 & \multicolumn{2}{c}{oracle (lowest error)} & \multicolumn{2}{c}{router, graded} & router, binary & random & median schedule\\
\cmidrule(lr){2-3}\cmidrule(lr){4-5}
task & blocks & vs.\ frontier & blocks & vs.\ frontier & vs.\ frontier & vs.\ frontier & vs.\ frontier\\
\midrule
acrobot-swingup & 11.1 & $0.72$ $[0.67, 0.76]$ & 7.2 & $0.95$ $[0.91, 1.00]$ & $1.05$ $[1.00, 1.10]$ & $1.01$ $[0.96, 1.05]$ & $1.00$ $[1.00, 1.00]$\\
cartpole-balance & 10.6 & $0.63$ $[0.58, 0.69]$ & 6.6 & $0.99$ $[0.93, 1.04]$ & $0.87$ $[0.76, 0.96]$ & $0.88$ $[0.81, 0.95]$ & $1.01$ $[1.00, 1.02]$\\
reacher-easy & 10.2 & $0.66$ $[0.61, 0.71]$ & 6.5 & $1.03$ $[0.96, 1.08]$ & $1.27$ $[1.16, 1.39]$ & $1.33$ $[1.23, 1.44]$ & $0.98$ $[0.94, 1.01]$\\
walker-walk & 7.4 & $0.77$ $[0.77, 0.78]$ & 6.6 & $0.99$ $[0.98, 1.00]$ & $1.29$ $[1.26, 1.32]$ & $1.29$ $[1.27, 1.32]$ & $0.97$ $[0.96, 0.98]$\\
finger-turn\_hard & 10.1 & $0.75$ $[0.71, 0.80]$ & 6.9 & $1.01$ $[0.98, 1.05]$ & $0.99$ $[0.91, 1.07]$ & $1.08$ $[1.04, 1.11]$ & $1.04$ $[1.01, 1.07]$\\
hopper-stand & 7.7 & $0.65$ $[0.65, 0.66]$ & 6.6 & $1.02$ $[1.00, 1.03]$ & $1.09$ $[1.07, 1.10]$ & $1.05$ $[1.03, 1.06]$ & $1.00$ $[0.99, 1.02]$\\
humanoid-run & 5.9 & $0.90$ $[0.90, 0.91]$ & 6.7 & $1.01$ $[1.00, 1.01]$ & $1.15$ $[1.14, 1.16]$ & $1.14$ $[1.14, 1.15]$ & $1.00$ $[0.99, 1.00]$\\
quadruped-walk & 5.7 & $0.82$ $[0.82, 0.83]$ & 6.8 & $1.05$ $[1.04, 1.06]$ & $1.07$ $[1.06, 1.08]$ & $1.07$ $[1.06, 1.09]$ & $1.01$ $[1.00, 1.02]$\\
cheetah-run & 7.5 & $0.68$ $[0.66, 0.70]$ & 6.7 & $1.04$ $[1.02, 1.07]$ & $1.23$ $[1.19, 1.27]$ & $1.24$ $[1.20, 1.29]$ & $1.08$ $[1.06, 1.10]$\\
\bottomrule
\end{tabular}}
\caption{\textbf{Routing at equal compute} ($K{=}4$, eight-seed means, bootstrap 95\% CIs). Each
``vs.\ frontier'' entry is the policy's rollout error over the best fixed-depth-mixture error at no more
than the policy's own average blocks per step; below 1 the policy beats every fixed-depth mixture.
Binary router at $\tau{=}0.5$; random routing with probability $0.5$ of the deep exit; schedule is the
oracle's median depth at each step index, applied to every window.}
\label{tab:routing}
\end{table}

\paragraph{Per-step choice has room to help.} The lowest-error oracle beats the fixed-depth
frontier on all nine tasks, by 10--37\% at its own budget of $5.7$--$11.1$ blocks per step, with every
confidence interval below 1 (Table~\ref{tab:routing}). The steps differ: along the full-depth
trajectory, some shallower exit beats depth 12 on 40--91\% of steps, including on every intrinsic
task (Appendix~\ref{app:routing}, Table~\ref{tab:prefs}). So a task where depth helps on average can
still contain many steps where it does not, and the reverse. This is the point one reviewer made
against our earlier framing, and it holds: $\rhov$ tests whether depth helps \emph{on average}, not
whether routing can help. The oracle also beats the schedule frontier, which a policy can reach
from the step index alone: by 8--28\% ($0.72$--$0.92$, every confidence interval below 1), even when
the schedules are chosen on the evaluation windows themselves (Table~\ref{tab:learned}). So its gain
depends on the state, not only on when in the rollout depth is spent. Two caveats keep the oracle from
being a target. It sees the true next latent, and it chooses among six predictions with hindsight, so
part of its gain is selecting whichever exit happened to land closest rather than spending compute
where it is needed. It bounds what a router with one-step lookahead could reach; it is not
achievable.

\paragraph{The difficulty router captures none of it.} The graded router runs $6.5$--$7.2$ blocks per
step and lands within 5\% of the frontier on every task ($0.95$--$1.05$). Its confidence interval is
entirely below 1 on one task, acrobot (mean $0.95$, upper bound $0.997$), and entirely above 1 on four, including all three where
a shallower depth is best. The binary router at $\tau{=}0.5$ does worse, $0.87$--$1.29$ of the
frontier, and on eight of nine tasks its confidence interval overlaps that of random routing at the
same probability. Applying the oracle's median depth at each step index as a schedule lands within 3\% of the
fixed-depth frontier on seven tasks, 4\% worse on finger and 8\% worse on cheetah; the searched
schedules below do better. One more
regularity: random routing between depths 2 and 12 is \emph{worse} than the frontier on seven of
nine tasks, so alternating between those two exits inside one rollout costs accuracy relative to
using either throughout. We have not tested why.

\paragraph{Why the router fails.} The difficulty head is trained to predict how large the full-depth
error will be, and it does: Spearman correlation $0.12$--$0.67$ with the realised full-depth error
(Table~\ref{tab:prefs}). It does not predict which exit is better. Its correlation with the gap
between depth-2 and depth-12 error is $-0.07$ to $0.10$ on the three tasks where a shallower depth is
best and $0.06$--$0.53$ elsewhere. A hard step for the full stack is not the same thing as a step the
full stack handles better than a shallow exit, and routing needs the second.

\begin{table}[tb]
\centering
\small
\setlength{\tabcolsep}{3pt}
\resizebox{\linewidth}{!}{%
\begin{tabular}{lccccccc}
\toprule
 & \multicolumn{3}{c}{exit-error router} & \multicolumn{2}{c}{oracle (lowest error)} & end-to-end & top-1\\
\cmidrule(lr){2-4}\cmidrule(lr){5-6}
task & vs.\ fixed & blocks & vs.\ schedule & vs.\ fixed & vs.\ schedule & vs.\ fixed & (base), \%\\
\midrule
acrobot-swingup & $0.94$ $[0.91, 0.97]$ & 2.8 & $1.30$ $[1.26, 1.35]$ & $0.72$ $[0.67, 0.76]$ & $0.89$ $[0.84, 0.93]$ & $1.03$ $[1.00, 1.07]$ & 60 (60)\\
cartpole-balance & $1.03$ $[0.95, 1.13]$ & 10.5 & $1.20$ $[1.04, 1.46]$ & $0.63$ $[0.58, 0.69]$ & $0.85$ $[0.79, 0.92]$ & $1.09$ $[1.04, 1.14]$ & 54 (55)\\
reacher-easy & $0.89$ $[0.85, 0.93]$ & 3.3 & $1.02$ $[0.98, 1.06]$ & $0.66$ $[0.61, 0.71]$ & $0.83$ $[0.79, 0.87]$ & $1.01$ $[0.99, 1.04]$ & 44 (45)\\
walker-walk & $0.88$ $[0.87, 0.89]$ & 2.9 & $1.02$ $[1.01, 1.03]$ & $0.77$ $[0.77, 0.78]$ & $0.83$ $[0.82, 0.85]$ & $1.21$ $[1.16, 1.26]$ & 19 (22)\\
finger-turn\_hard & $1.00$ $[0.99, 1.00]$ & 2.0 & $1.00$ $[0.99, 1.01]$ & $0.75$ $[0.71, 0.80]$ & $0.80$ $[0.76, 0.85]$ & $1.10$ $[1.04, 1.19]$ & 47 (48)\\
hopper-stand & $0.92$ $[0.91, 0.93]$ & 3.1 & $1.01$ $[1.00, 1.02]$ & $0.65$ $[0.65, 0.66]$ & $0.73$ $[0.72, 0.75]$ & $1.06$ $[1.04, 1.09]$ & 24 (28)\\
humanoid-run & $0.96$ $[0.95, 0.96]$ & 2.8 & $1.01$ $[1.00, 1.01]$ & $0.90$ $[0.90, 0.91]$ & $0.92$ $[0.91, 0.93]$ & $1.12$ $[1.11, 1.13]$ & 21 (26)\\
quadruped-walk & $0.99$ $[0.99, 1.00]$ & 2.9 & $1.00$ $[1.00, 1.00]$ & $0.82$ $[0.82, 0.83]$ & $0.84$ $[0.83, 0.84]$ & $1.05$ $[1.04, 1.06]$ & 20 (30)\\
cheetah-run & $0.95$ $[0.94, 0.95]$ & 3.2 & $1.00$ $[0.99, 1.02]$ & $0.68$ $[0.66, 0.70]$ & $0.72$ $[0.70, 0.74]$ & $1.18$ $[1.13, 1.25]$ & 17 (25)\\
\bottomrule
\end{tabular}}
\caption{\textbf{Routers trained for the routing decision} ($K{=}4$, eight seeds, bootstrap 95\%
CIs). ``vs.\ fixed'': error over the fixed-depth frontier at the policy's own blocks per step.
``vs.\ schedule'': error over the schedule frontier at the same budget, with schedules chosen on
validation windows for the router (fair to the router) and on the evaluation windows for the oracle
(favouring the schedules). The exit-error router's $\lambda$ is chosen for each seed from the task's
other seeds. Top-1: how often the router's predicted best exit is the true best at one step, with the
rate of always predicting the most common best exit in parentheses.}
\label{tab:learned}
\end{table}

\paragraph{A router trained for the decision.} The difficulty head was never trained to choose an
exit, so we trained routers that are (Section~\ref{sec:setup}). The exit-error router, with $\lambda$
chosen on held-out seeds, beats the fixed-depth frontier on seven tasks at $2.8$--$3.3$ blocks per step:
by 4--12\% on six ($0.88$--$0.96$) and by 1\% on quadruped (Table~\ref{tab:learned}). It does not on
finger, where the chosen $\lambda$ sends almost every step to depth 2, or on cartpole. The end-to-end
router is no better than the frontier on any task ($1.01$--$1.21$; significantly worse on seven).

\paragraph{Its gain is timing, not state.} Against the schedule frontier the exit-error router's
advantage disappears. With schedules chosen on validation windows it is within 2\% of the best
schedule on seven tasks ($1.00$--$1.02$) and worse on acrobot ($1.30$) and cartpole ($1.20$); it is below the
schedule frontier on none. The router sees the step index, and what it learned is when in the
rollout to spend depth, which a fixed schedule does as well. It did not learn which states need depth.
At one step its predicted best exit is right no more often than always guessing the most common best
exit, on every task (top-1 $17$--$60\%$ against $22$--$60\%$). Its predictions of each exit's error
relative to full depth explain little of the variance on held-out windows (mean $R^2$ over exits
$0.00$--$0.37$).

\paragraph{What this shows and what it does not.} Choosing depth per step has room to help on every
task, beyond anything fixed depths or schedules reach, and none of the three routers we tested finds
it: the difficulty head is at the fixed-depth frontier, the exit-error router at the schedule
frontier, and the end-to-end router above it. The simplest reading is that the exit that will be most
accurate is close to unpredictable from $(z_t, a_t)$ in these models, so the oracle's advantage is
largely hindsight: it picks whichever of six noisy predictions happened to land nearest the truth.
We have not ruled out that a larger router, more data, or a different input (for example the exits'
own disagreement) would find it. The fixed-depth comparisons hold on the larger MLP, the transformer,
the pixel models and the models trained on competent-policy data (Appendix~\ref{app:variants}), even
where the fixed-depth effects themselves shrink.

\section{Exit supervision and the cheetah inversion}\label{sec:catch22}
Is cheetah's inversion a property of its dynamics or of how the exits are trained? Our backbone applies
deep supervision and distillation to every shallow exit at every rollout step, so each exit is trained to be a usable one-step operator anywhere in a rollout.
We hypothesised that this per-step training is what makes the shallow exits better rollout
operators than the full stack.

\paragraph{Ablation.} A single training flag, \texttt{deep\_sup\_first\_step\_only}, computes the
shallow-exit terms (deep supervision and distillation) only at the first rollout step and leaves the
full-depth objective unchanged at all steps. We retrained cheetah, quadruped (the other inversion
task) and walker (an intrinsic control) at $K{=}4$ with the flag set. The pre-registered test,
\textbf{Gate B}, required cheetah's mean $\rhov$ to rise by at least $0.15$ (Table~\ref{tab:catch22}).

\begin{table}[tb]
\centering
\small
\setlength{\tabcolsep}{4pt}
\begin{tabular}{lcccc}
\toprule
task & standard $K{=}4$ & first-step-only $K{=}4$ & $\Delta$ & $n$\\
\midrule
cheetah-run & 0.869 (0.81--0.91) & 1.147 (1.06--1.26) & $+0.28$ & 8\\
quadruped-walk & 0.893 (0.88--0.90) & 0.854 (0.85--0.86) & $-0.04$ & 8/3\\
walker-walk & 1.950 (1.89--2.00) & 2.064 (2.01--2.13) & $+0.11$ & 8/3\\
\bottomrule
\end{tabular}
\caption{\textbf{Gate B}: $\rhov$ under standard and first-step-only exit supervision, means with
per-seed min--max ($n{=}8$ for standard arms and the cheetah ablation, $n{=}3$ for the other
ablations). Cheetah's rise of $0.28$ passes the pre-registered threshold of $0.15$, and the two
cheetah distributions do not overlap (standard $\le0.91$, ablated $\ge1.06$; two standard seeds, at
$0.905$ and $0.909$, are just above the $0.90$ inversion threshold). Quadruped's inversion is not
removed, and walker stays strongly intrinsic.}
\label{tab:catch22}
\end{table}

\begin{table}[tb]
\centering
\small
\setlength{\tabcolsep}{4pt}
\begin{tabular}{llccc}
\toprule
task & quantity & standard & first-step-only & paired change\\
\midrule
cheetah-run ($n{=}8$) & one-step error, depth 2 & 0.062 & 0.062 & $-0.3\%$ $[-3.1, +2.3]$\\
cheetah-run ($n{=}8$) & one-step error, depth 12 & 0.052 & 0.060 & $+16\%$ $[+10, +21]$\\
cheetah-run ($n{=}8$) & rollout error, depth 2 & 0.142 & 0.166 & $+16\%$ $[+13, +20]$\\
cheetah-run ($n{=}8$) & rollout error, depth 12 & 0.164 & 0.145 & $-12\%$ $[-15, -8]$\\
cheetah-run ($n{=}8$) & ratio at one step, $\rho_1$ & 1.207 & 1.040 & $-0.167$ $[-0.218, -0.118]$\\
cheetah-run ($n{=}8$) & ratio after rollout, $\rho$ & 0.869 & 1.147 & $+0.278$ $[+0.237, +0.329]$\\
\midrule
quadruped-walk ($n{=}3$) & one-step error, depth 2 & 0.206 & 0.218 & $+6\%$ $[+5, +8]$\\
quadruped-walk ($n{=}3$) & one-step error, depth 12 & 0.229 & 0.259 & $+13\%$ $[+10, +17]$\\
quadruped-walk ($n{=}3$) & rollout error, depth 2 & 0.317 & 0.325 & $+3\%$ $[+0, +4]$\\
quadruped-walk ($n{=}3$) & rollout error, depth 12 & 0.355 & 0.381 & $+7\%$ $[+6, +10]$\\
quadruped-walk ($n{=}3$) & ratio at one step, $\rho_1$ & 0.898 & 0.845 & $-0.054$ $[-0.070, -0.040]$\\
quadruped-walk ($n{=}3$) & ratio after rollout, $\rho$ & 0.892 & 0.854 & $-0.037$ $[-0.048, -0.019]$\\
\midrule
walker-walk ($n{=}3$) & one-step error, depth 2 & 0.102 & 0.089 & $-13\%$ $[-15, -11]$\\
walker-walk ($n{=}3$) & one-step error, depth 12 & 0.042 & 0.033 & $-20\%$ $[-21, -19]$\\
walker-walk ($n{=}3$) & rollout error, depth 2 & 0.354 & 0.330 & $-7\%$ $[-8, -5]$\\
walker-walk ($n{=}3$) & rollout error, depth 12 & 0.181 & 0.160 & $-12\%$ $[-16, -7]$\\
walker-walk ($n{=}3$) & ratio at one step, $\rho_1$ & 2.452 & 2.666 & $+0.214$ $[+0.201, +0.241]$\\
walker-walk ($n{=}3$) & ratio after rollout, $\rho$ & 1.959 & 2.064 & $+0.105$ $[-0.009, +0.194]$\\
\bottomrule
\end{tabular}
\caption{\textbf{What the ablation changes} ($K{=}4$; errors are seed means, paired changes are
means over seeds with bootstrap 95\% CIs; walker and quadruped use the three seeds that have both
arms). One-step error starts each step from the encoded true state; rollout error is averaged over a
ten-step rollout.}
\label{tab:gateb}
\end{table}

\paragraph{What moves.} Table~\ref{tab:gateb} separates one-step from rollout error at both ends of
the network. Under standard training, cheetah's depth-2 exit is the worse one-step predictor
($1.21\times$ the full-depth error) and the better rollout operator ($0.87\times$): its error grows by
a factor of $2.3$ over ten steps, against $3.2$ for the full stack. The ablation leaves depth 2's
one-step error unchanged ($-0.3\%$, CI $[-3.1, +2.3]$) and raises its rollout error by $16\%$; its growth
factor rises to $2.7$. For the shallow exit, then, the ablation acts on how errors compound and not
on one-step accuracy, which is the specific concern that the exits were simply ``supervised less''.
The full-depth path changes too. Its one-step error rises $16\%$ while its rollout error falls
$12\%$, and its growth factor drops to $2.4$. The exit losses share blocks with the full stack, so
changing them changes the full model, and $\rhov$ rises from both ends. This ablation therefore shows
that cheetah's inversion is a training effect, since the same architecture and data stop inverting
when only the exit-supervision schedule changes, but it does not isolate the cause to the shallow
exit.

\paragraph{A supervision-matched control.} One reviewer read the ablation as simply supervising the
exits less. It does not reduce the total exit-loss weight: the exit terms are averaged over rollout
steps, so under the flag the step-0 term carries the whole weight that four steps share in standard
training. What the flag changes is two things at once: each sample has one supervised exit prediction
instead of four, and that prediction always starts from an encoded true state, never from a composed
one. To separate them we added a control that supervises the exits at one rollout step per sample,
drawn uniformly, with the same total weight. It has the same number of supervised predictions as the
ablation, but three times in four they start from a composed latent. We retrained all three arms of
cheetah at $K{=}4$ for seeds 0--2 on different hardware from the main campaign; the standard and
first-step-only arms reproduce the original runs of the same seeds to within $0.04$ and $0.05$ in
$\rhov$.

\begin{table}[H]
\centering
\small
\setlength{\tabcolsep}{3pt}
\resizebox{\linewidth}{!}{%
\begin{tabular}{lcccccccc}
\toprule
 & & & \multicolumn{2}{c}{one-step error} & \multicolumn{2}{c}{rollout error} & \multicolumn{2}{c}{growth}\\
\cmidrule(lr){4-5}\cmidrule(lr){6-7}\cmidrule(lr){8-9}
exit supervision & $\rhov_1$ & $\rhov$ & depth 2 & depth 12 & depth 2 & depth 12 & depth 2 & depth 12\\
\midrule
standard (every step) & $1.16$ $[1.07, 1.26]$ & $0.85$ $[0.83, 0.89]$ & 0.060 & 0.052 & 0.140 & 0.165 & 2.33 & 3.17\\
first step only & $1.04$ $[1.03, 1.05]$ & $1.14$ $[1.11, 1.16]$ & 0.061 & 0.058 & 0.159 & 0.139 & 2.63 & 2.39\\
one random step & $1.04$ $[0.99, 1.12]$ & $0.91$ $[0.83, 0.96]$ & 0.066 & 0.064 & 0.143 & 0.159 & 2.18 & 2.50\\
\bottomrule
\end{tabular}}
\caption{\textbf{Supervision-matched control on cheetah} ($K{=}4$, three seeds per arm, means with
bootstrap 95\% CIs). All three arms have the same total exit-loss weight. ``Growth'' is rollout error
over one-step error.}
\label{tab:control}
\end{table}

The control keeps depth 2 the better rollout operator on every seed ($\rhov$ $0.83$--$0.96$, mean
$0.91$), while first-step-only supervision reverses that on every seed ($1.11$--$1.16$; Table~\ref{tab:control}). Paired
by seed, the control moves $\rhov$ by $0.00$ to $+0.08$ from the standard arm and the ablation by
$+0.28$ to $+0.31$, so the control keeps about $80\%$ of the inversion. The one-step picture
separates the two from the rollout picture: both the control and the ablation bring the one-step ratio
down to about $1.04$, but only the ablation raises depth 2's rollout error ($+14\%$ against $+2\%$ for
the control) and its error growth ($2.63$ against $2.18$, standard $2.33$). So the inversion depends on
the exits being supervised from composed latents, not on how much or how often they are supervised.
Two limits remain. The control's mean sits at the $0.90$ threshold we use to label inversion, so it
weakens the inversion a little as well as keeping most of it. And in both arms that supervise
composed inputs, the exit losses also send gradient back through the earlier full-depth steps that
produced those inputs; the control does not separate the exit's own training from that effect on the
full path.

\paragraph{The other two tasks.} Quadruped's inversion is a different kind. Its depth-2 exit already
beats the full stack at one step, by the same margin it shows after rollout ($\rhov_1{=}0.89$,
$\rhov{=}0.89$; Table~\ref{tab:depth}), so composition plays no part in it: it is the one-step
overthinking effect, not a compounding effect. That is why an intervention aimed at composition
leaves it in place. Under the ablation every quadruped error rises slightly and both ratios fall a
little together ($0.90\to0.85$ at one step, $0.89\to0.85$ after rollout). Why the full stack
predicts worse than its first two blocks on this task we have not established. On walker every error falls and
depth keeps its large advantage. A third observation applies to cheetah only: standard per-step exit
supervision helps the full stack's one-step accuracy but costs it rollout accuracy, so on this task
the auxiliary exit losses make the full model a better single-step predictor and a worse rollout
operator. Section~\ref{sec:cautions} adds one more qualification: with the same loss, cheetah does
not invert when trained on data from a competent policy, so the effect depends on the training data
as well as the loss.

\section{What the regime depends on}\label{sec:cautions}
The taxonomy is measured in one configuration (latent-space error, $H{=}10$, MLP encoder,
random-policy data); we now vary each axis. The \emph{fine} label is condition-dependent, but a
stable invariant survives: across four measurement conditions (latent-$H\{10,30,50\}$, probe-$H10$)
fine-label agreement with the canonical read is $0.67$, while the
\emph{dichotomy}---does depth help on average or not---agrees on \textbf{7 of 9 tasks} (acrobot has
no probe measurement and agrees in the other three conditions). The two exceptions, cartpole and
reacher, read flat only under the probe, which we attribute to a measurement floor on
near-perfectly-modeled tasks.

\paragraph{Metric space.} $\rhov$ is computed in latent space, the default for JEPA-style models. As
a control we fit a linear probe from frozen latents to observations (held-out $R^2\in[0.95,0.98]$)
and recompute the depth curves as observation-space error, having pre-registered a kill criterion: if
any task's label changes between spaces, the taxonomy is metric-dependent and must be reported as
such. \textbf{It fired.} Humanoid is \emph{flat} in latent space ($\rhov^{K4}=0.96$) but
\emph{inverted} in observation space ($0.78$)---the shallow exits decode to better observations than
the full stack, an advantage the low-variance latent geometry masks (Figure~\ref{fig:metric}); cheetah and walker are
stable. On this task, then, latent prediction error, the standard world-model metric, misstates
whether depth buys quality. One task does not show how often this happens, but it shows that the
latent metric alone can mislead.

\paragraph{Horizon, encoder, and backbone.} Three further axes move the label; we summarize them here
and give the full analysis with figures in Appendix~\ref{app:axes}. Each perturbation arm is three
seeds and is read against its \emph{matched} three-seed baseline, so the reference values below can
differ slightly from Table~\ref{tab:taxonomy}'s eight-seed means. \emph{Horizon}: cheetah's
inversion is \emph{short-horizon}---at $H\in\{30,50\}$ the shallow exits' per-step errors compound and
depth's advantage recovers ($\rhov^{K4}:0.85\to0.98\to1.07$)---so short-horizon MPC and long-horizon
rollout can sit on opposite sides of $\rhov=1$. \emph{Encoder}: retrained from $64\times64$ pixels
with a CNN encoder, the two strongly intrinsic state tasks lose their depth advantage (at $K{=}1$,
cartpole $2.15\to0.85$ and walker $1.67\to0.87$; under the pre-registered rule all three pixel tasks
are flat), and not merely as a model-quality effect: \textbf{walker loses it at matched one-step
error} ($0.0043$ vs.\ $0.0044$, each in its own latent space). \emph{Backbone}: under a $2\times$-wider, $2\times$-deeper MLP and a
transformer predictor the pinned $\ge2/3$-dichotomy criterion confirms for both families, and
\textbf{cheetah's inversion reproduces in the transformer family} ($\rhov^{K4}=0.88$), so
the taxonomy's surprising half is not a residual-MLP artifact; under the larger MLP cheetah sits just
above the inversion threshold ($0.92$). Neither half of the taxonomy is fixed across these axes. The
inversion disappears at longer horizons and is lost or weakened with pixels ($K{=}4$: $0.91$) and with
competent-policy data ($1.06$), and the size of depth's advantage on intrinsic tasks shrinks with
pixels and larger backbones. What holds is narrower: cheetah never enters the intrinsic regime
($\rhov^{K1}<1.25$) under any horizon, encoder, backbone or data distribution we tried.

\paragraph{Training distribution.} The single largest lever is the data the model is trained on
(Section~\ref{sec:planner}): retraining cheetah and cartpole from competent planner-collected data
instead of random-policy data \emph{erases} both the inversion and the strong intrinsic tradeoff,
collapsing both toward flat---even though the loss is unchanged. This is $n{=}2$ tasks at a single
competent-data regime, so we do not claim every regime washes out on-policy. What the two cases do
establish is that $\rhov$ measured on random-exploration data cannot be read as a fixed property of
the task: at least sometimes the depth difference belongs to the broad exploration distribution and
closes on the narrow states a competent planner visits. Appendix~\ref{app:randomdata} argues why the
random-data regime is still the relevant one for the decisions this diagnostic informs.

\paragraph{A null mechanism.} We also asked whether contractivity explains the inversion. Measuring
exact Jacobian spectral norms and $50$-step latent-norm trajectories
(Appendix~\ref{app:contractivity}), latent norms are uniformly stable ($\approx1.0\times$) and
spectral norms do not separate shallow from deep in a regime-predictive way. We report this as a
null result: the ablation in Section~\ref{sec:catch22} points to training rather than a fixed-point
property of the maps.

\section{Is the regime predictable a priori?}\label{sec:predict}
A descriptive taxonomy is more useful if the regime can be anticipated before training a full model.
We correlate the nine tasks' $\rhov^{K4}$ against measurable task/model properties
(Appendix~\ref{app:correlates}). Three predictors stand out: \emph{observation dimensionality}
(Spearman $-0.79$), \emph{action dimensionality} ($-0.83$)---both known a priori from the task---and
\emph{full-depth rollout error} ($-0.78$). The flat and inversion tasks include the two
highest-dimensional ones (quadruped, humanoid) and are hard to model, while the strong intrinsic
tasks are low-dimensional and modeled almost perfectly
($e_{\mathrm{full}}\!\le\!4\times10^{-3}$); cheetah ($17$ obs) is the exception, lower-dimensional
than intrinsic walker ($24$). At $n{=}9$ with collinear predictors these are
descriptive associations, not a validated predictor---we deliberately avoid partial correlations or
significance claims at this sample size. They say that depth tends to pay on tasks the model fits almost perfectly and not on hard,
high-dimensional ones; they do not give a mechanism. The simplest story, that deep predictions are
noisier per step on hard tasks and that the noise compounds, does not fit our own data: on cheetah
the full stack is the more accurate one-step predictor ($\rhov_1{=}1.21$, Table~\ref{tab:depth}), and
longer rollouts restore depth's advantage rather than deepen the inversion
(Section~\ref{sec:cautions}). Section~\ref{sec:catch22} points instead to how the exits are trained.

\paragraph{An out-of-sample test.} To turn correlation into prediction we froze a logistic classifier
on $[\log\text{obs-dim}, \log(\text{act-dim}{+}1)]$ (fit on the nine in-sample tasks; $9/9$ correct),
committed its predictions for six \emph{held-out} tasks to git, and only then trained them. Two caveats. Its fitted weights are
$[{+}2.21,{-}7.50]$: the \emph{action}-dimensionality term dominates and the observation weight is
\emph{positive}---the opposite sign to the marginal Spearman above, since walker (intrinsic) and
cheetah (non-intrinsic) share an action dimension and differ only in observation dimension. The rule
is thus action-dimensionality-driven, not a confirmation of ``high-dimensional tasks invert''. And of
the held-out set only two tasks are informative: three (hopper-hop, walker-run, walker-stand) share physics,
observations and random-policy data with in-sample tasks, so they produce identical models and
identical $\rhov$ and are not new measurements, and cartpole-swingup has exactly in-sample cartpole-balance's dimensionality ($5$ obs, $1$ act), so a
dimensionality-only rule must reproduce its training label. On the two genuinely new tasks the rule is right once. \textbf{The extrapolation lands}: dog-walk---predicted
non-intrinsic at $P(\text{intrinsic}){=}0.00$ from a dimensionality ($223$ obs, $38$ act) far beyond
the training range---is empirically the strongest, most seed-consistent inversion we measured
($\rhov^{K4}=0.70$, all seeds, both $K$); by the weights above this call is driven by its $38$-dimensional
action space. The miss, fish-upright ($P{=}0.83$), is itself informative: it is the first empirical
instance of the \emph{created} regime ($\rhov^{K1}{=}1.12$, $\rhov^{K4}{=}1.68$)---the regime
Section~\ref{sec:negative}'s hypothesis predicted but did not find on its first three tasks. At
$n{=}2$ genuinely-novel tasks this is suggestive rather than a validated predictor, but the extreme
extrapolation is evidence that the regime is anticipatable rather than purely post-hoc.

\section{Fixed depths in closed-loop planning}\label{sec:planner}
$\rhov$ is measured from open-loop rollout error. If it is a meaningful diagnostic for embodied use
and not an artifact of that metric, its sign should predict whether \emph{planning} benefits from
depth. On each $K{=}4$ checkpoint we fit a ridge reward head on frozen latents and run a minimal CEM
planner ($256$ candidates, horizon $10$, 2 iterations) closed-loop on the true environment, at fixed
depth 2, fixed depth 12, and routed with the binary rule at $\tau{=}0.5$. Each condition is evaluated
on three $200$-step episodes per model seed. Every condition uses the same number of candidates, so
the depth-12 planner spends $5.6\times$ the predictor compute of the depth-2 planner: these
comparisons are at equal candidates, not equal compute. We pre-registered two predictions before collecting
any depth-vs-return data: \textbf{P1}---the sign of $\mathrm{return}(d_{12})-\mathrm{return}(d_2)$
matches the regime (positive on intrinsic, non-positive on inversion/flat); \textbf{P2}---routed
planning reaches $\ge0.9\times$ the full-depth return at $\ge25\%$ block savings on intrinsic tasks.
A task is \emph{viable} iff its reward $R^2\ge0.3$ and the full-depth planner beats random.

\paragraph{P1 holds where the regime and the reward model are robust (Figure~\ref{fig:planner}).} The
``deep helps'' side is confirmed on \emph{two} intrinsic tasks: cartpole ($R^2{=}0.99$), where deep
planning beats shallow by $+32$ return, and acrobot ($R^2\approx0.70$), on both viable seeds. The
``deep hurts'' side shows up on cheetah: on \textbf{cheetah (robust inversion) shallow planning
beats deep} ($7.24$ vs $6.58$ mean return over $n{=}8$ seeds, shallow winning on $6/8$)---with this
planner the rollout-error inversion reproduces in closed-loop control. Of the six tasks with a usable reward model, P1
holds on four and misses on two---both already flagged by the taxonomy: quadruped is the borderline,
seed-unstable inversion, and walker's reward model is weak ($R^2{=}0.55$). Hopper and reacher are unviable: hopper's reward head misses the $R^2\ge0.3$ bar ($0.16$), and on reacher, whose reward is sparse, the full-depth planner never beats random;
finger-turn\_hard was not run in the planner. Episodes are $200$ steps, so cheetah's $\approx7$ is far below a trained agent's:
P1's sign is measured in this weak-planner regime. The validation is therefore \emph{partial but two-sided}, and degrades
gracefully exactly where the regime signal or the reward model is weak.

\paragraph{P2 fails at the fixed threshold; a calibrated router partly recovers it.} At the default
threshold ($0.5$), routed planning on cartpole saves $70\%$ of predictor blocks but at $0.82\times$
the full-depth return---below the bar (P2, pre-registered, fails). We then pre-registered a
\emph{calibrated} router (P2$'$): sweep the threshold on calibration episodes, pick the
highest-savings threshold meeting $0.9\times$ there, and evaluate it frozen on a \emph{disjoint} set
(no shared seeds, so this is not post-hoc tuning). It clears the bar on \textbf{acrobot for all three
seeds}---$1.01$--$1.09\times$ full depth at $83\%$ block savings, i.e.\ routing is essentially
free---and is borderline on cartpole (mean $0.84\times$). So with this planner the sign of $\rhov$
tracks whether depth helps on the robust tasks, and calibration turns that into a favorable operating
point on some intrinsic tasks; the stronger planner below shows that the sign itself can change with
the planner.

\paragraph{The control-level effect depends on the planner and on the data.} \emph{(i) Planner
strength.} Under a stronger planner (MLP reward head, three on-policy aggregation rounds, $512$
candidates, three CEM iterations with warm-starting, $500$-step episodes; $R^2$ $0.84$--$1.00$), cheetah's shallow-beats-deep \emph{reverses}. In that setting the difficulty-routed planner
beat both fixed depths on the two inversion tasks at equal candidates (quadruped $1.12\times$ the
better of the two at $21\%$ fewer blocks, cheetah $1.05\times$ at $14\%$), while full depth won on the
intrinsic tasks. We test that result at equal compute below. \emph{(ii) Training distribution.} Retraining cheetah
and cartpole from competent-planner data rather than random-policy data \emph{collapses both regimes
toward flat} (inversion $0.85\to1.06$; intrinsic $2.93\to1.00$) even though the loss is
unchanged. With the planner-strength result, this says that \emph{whether depth helps is a property
of the operating regime, not of the task alone}.

\paragraph{At equal compute, routing matches a fixed middle depth.} The routed planner above mixes
depths 2 and 12, and the comparison gave it the same number of candidates as each fixed depth, so it
was compared with the two endpoints only and the depth-12 planner spent more compute. We reran it on
the two inversion tasks against every fixed exit at equal predictor compute: a fixed depth $d$
planner gets $N_d=\mathrm{round}(512\,\bar B/d)$ candidates and $N_d/8$ elites, where $\bar B$ is the
routed planner's measured blocks per step, which matches compute to within $0.1\%$. Everything else
is the stronger planner above, including its viability rule, with three model seeds and three
$500$-step episodes each.

\begin{table}[H]
\centering
\small
\setlength{\tabcolsep}{3pt}
\resizebox{\linewidth}{!}{%
\begin{tabular}{lccccccccccc}
\toprule
 & & \multicolumn{2}{c}{routed} & \multicolumn{6}{c}{fixed depth, same compute as routed} & best & routed /\\
\cmidrule(lr){3-4}\cmidrule(lr){5-10}
task & seeds & blocks & return & 2 & 4 & 6 & 8 & 10 & 12 & depth & best fixed\\
\midrule
cheetah-run & 3 & 9.8 & 34.9 & 26.4 & 31.5 & 35.6 & 35.1 & 35.2 & 32.0 & 6 & 0.98\\
quadruped-walk & 2 & 9.8 & 169.1 & 109.6 & 171.3 & 167.5 & 165.8 & 154.1 & 167.6 & 4 & 0.99\\
\bottomrule
\end{tabular}}
\caption{\textbf{Closed-loop return at equal predictor compute} (mean over viable model seeds, three
episodes each). Each fixed depth gets as many candidates as the routed planner's compute allows. Per
seed curves are in Figure~\ref{fig:planner_ec}.}
\label{tab:planner_ec}
\end{table}

\begin{figure}[H]
\centering
\includegraphics[width=0.62\linewidth]{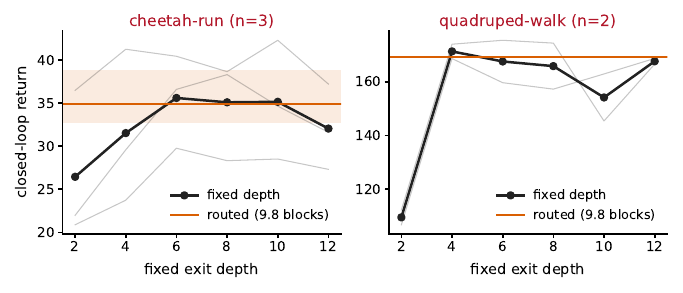}
\caption{\textbf{Fixed depth against routing at equal compute.} Black: return of each fixed exit
given the routed planner's compute (thin lines are model seeds). Orange: the routed planner (band:
range over seeds).}
\label{fig:planner_ec}
\end{figure}

The routed planner beats both endpoints at equal compute, as before: on cheetah it returns $34.9$
against $32.0$ at depth 12 and $26.4$ at depth 2. But a fixed middle depth does as well
(Table~\ref{tab:planner_ec}). On cheetah depths 6, 8 and 10 return $35.1$--$35.6$ and routing reaches
$0.98$ of the best; on quadruped depth 4 returns $171.3$ and routing $0.99$ of it. Routing beats every
fixed depth on one of three cheetah seeds and one of two quadruped seeds. This is the open-loop
picture of Section~\ref{sec:routing} in closed loop: what the difficulty router gains over the
endpoints, a fixed intermediate depth also gains.

Two disclosures. First, the reward heads of the original strong-planner runs for cheetah were
overwritten by a later experiment, so we refit all heads with the original procedure ($R^2$
$0.90$--$0.95$). At equal candidates, depth-12 returns reproduce the original runs on three seeds
(within $2\%$) and not on two cheetah seeds ($33.9\to27.5$, $50.7\to33.4$), and quadruped seed 0
fails the viability rule with the refit head (depth 12 returns $64.5$ against $99.9$ for random; the
original run passed narrowly, $103.7$). Table~\ref{tab:planner_repro} lists every seed. Second,
quadruped returns are set mostly by the episode, not the planner: on one evaluation episode every
condition scores $4.6$--$8.7$; on another the random policy already scores $\approx249$ and planners score
$69$--$267$; only the third separates the planners ($61$--$262$, random $30$). The quadruped
comparison therefore rests on about one informative episode per seed. Cheetah has no such
structure: every condition scores $17$--$45$ on every episode and random scores below $3$.

\section{Discussion and limitations}
\paragraph{What this means for adaptive-compute world models.} Two of the questions a designer of
such a model faces have separate answers here. Whether depth helps on average depends on the task,
the training objective and the data: on three of nine tasks it does not, and composition moves the
answer away from what one-step error suggests. Whether per-step choice can help has a consistent
answer in hindsight: on every task, an oracle that sees the next latent beats every fixed depth and
every schedule we found at equal compute. Whether a router can reach that gain from what it sees has,
so far, a negative answer. The difficulty head stays at the fixed-depth frontier; a router trained to
predict each exit's error learns when in a rollout to spend depth, which a fixed schedule does as
well; an end-to-end router does worse; and in closed loop the difficulty-routed planner matches a
fixed middle depth. For evaluation, three practices follow. Report the full depth curve, at one step
and after rollout, rather than one ratio. Compare any router at equal compute against every fixed
depth and against step-index schedules, not only against the shallowest and deepest exit: our routed
planner beats both endpoints but not depth 6 on cheetah or depth 4 on quadruped, and our best learned
router beats every mixture of fixed depths but no schedule. And report whether the router's predicted
best exit beats always predicting the most common one; ours did not.

\paragraph{Limitations.} The main campaign is state-based DMC with a small residual-MLP predictor.
That was deliberate, since the encoder is itself one of the factors that move the results
(Section~\ref{sec:cautions}), but it leaves a gap to pixel- and sensor-driven systems, and one
reviewer noted that larger models and better data weaken the effects we report. For the fixed-depth
effects our scaling, transformer and competent-data runs agree; the routing results do not weaken
(Appendix~\ref{app:variants}). The labels are specific to the configuration they were
measured in: ``cheetah is an inversion task'' means it inverts under random-policy, state-space
training at $H{=}10$. The thresholds are reasonable but not derived; we report raw ratios. The
supervision-matched control on cheetah has three seeds, trained on different hardware from the main
campaign, does not separate the exits' own training from the gradient their losses send through the
full path, and does not explain quadruped. The routing oracle is greedy and uses
hindsight, so its gains are an upper bound and partly reflect selection among six predictions. We
tested three router designs, all small MLPs on $(z_t, a_t)$ and the step index; a larger router or
other inputs might do better. The schedule search covers two-phase and greedy schedules, not all
$6^{10}$ depth sequences. The equal-compute planner runs cover two tasks with refit reward heads, three
model seeds (two viable on quadruped) and three episodes each, and quadruped's returns turn on about
one episode per seed. The out-of-sample predictor rests on two genuinely new tasks. The fixed-depth campaign was
pre-registered (Appendix~\ref{app:prereg}); the routing analyses, the equal-compute planner runs, the
ablation control and the one-step decomposition were not.

\paragraph{Conclusion.} In this testbed, more predictor depth does not reliably give a better rollout,
and composition can move the depth--error relation in either direction relative to one step. In
hindsight, choosing depth per step beats every fixed depth and every step-index schedule we found, on
every task we measured. None of the routers we trained captures that: the best learns when in a
rollout to spend depth, which a fixed schedule does as well, and in closed loop the routed planner
matches a fixed middle depth. Whether the per-step gain is reachable from the state, or mostly
reflects hindsight selection among noisy exits, is the open question. A claim that it is reachable
should be tested against the best fixed depth and the best schedule at equal compute.

\bibliographystyle{plainnat}
\bibliography{references}

\begin{thebibliography}{44}
\providecommand{\natexlab}[1]{#1}
\providecommand{\url}[1]{\texttt{#1}}
\expandafter\ifx\csname urlstyle\endcsname\relax
  \providecommand{\doi}[1]{doi: #1}\else
  \providecommand{\doi}{doi: \begingroup \urlstyle{rm}\Url}\fi

\bibitem[Agarwal et~al.(2021)Agarwal, Schwarzer, Castro, Courville, and
  Bellemare]{agarwal2021rliable}
Rishabh Agarwal, Max Schwarzer, Pablo~Samuel Castro, Aaron Courville, and
  Marc~G. Bellemare.
\newblock Deep reinforcement learning at the edge of the statistical precipice.
\newblock In \emph{Advances in Neural Information Processing Systems
  (NeurIPS)}, 2021.

\bibitem[Asadi et~al.(2018)Asadi, Misra, and Littman]{asadi2018lipschitz}
Kavosh Asadi, Dipendra Misra, and Michael~L. Littman.
\newblock Lipschitz continuity in model-based reinforcement learning.
\newblock In \emph{International Conference on Machine Learning (ICML)}, 2018.

\bibitem[Assran et~al.(2023)Assran, Duval, Misra, Bojanowski, Vincent, Rabbat,
  LeCun, and Ballas]{assran2023ijepa}
Mahmoud Assran, Quentin Duval, Ishan Misra, Piotr Bojanowski, Pascal Vincent,
  Michael Rabbat, Yann LeCun, and Nicolas Ballas.
\newblock Self-supervised learning from images with a joint-embedding
  predictive architecture.
\newblock In \emph{IEEE Conference on Computer Vision and Pattern Recognition
  (CVPR)}, 2023.

\bibitem[Banino et~al.(2021)Banino, Balaguer, and
  Blundell]{banino2021pondernet}
Andrea Banino, Jan Balaguer, and Charles Blundell.
\newblock {PonderNet}: Learning to ponder.
\newblock \emph{arXiv preprint arXiv:2107.05407}, 2021.

\bibitem[Bardes et~al.(2022)Bardes, Ponce, and LeCun]{bardes2022vicreg}
Adrien Bardes, Jean Ponce, and Yann LeCun.
\newblock {VICReg}: Variance-invariance-covariance regularization for
  self-supervised learning.
\newblock In \emph{International Conference on Learning Representations
  (ICLR)}, 2022.

\bibitem[Bardes et~al.(2024)Bardes, Garrido, Ponce, Chen, Rabbat, LeCun,
  Assran, and Ballas]{bardes2024vjepa}
Adrien Bardes, Quentin Garrido, Jean Ponce, Xinlei Chen, Michael Rabbat, Yann
  LeCun, Mahmoud Assran, and Nicolas Ballas.
\newblock Revisiting feature prediction for learning visual representations
  from video.
\newblock \emph{arXiv preprint arXiv:2404.08471}, 2024.

\bibitem[Bolukbasi et~al.(2017)Bolukbasi, Wang, Dekel, and
  Saligrama]{bolukbasi2017adaptive}
Tolga Bolukbasi, Joseph Wang, Ofer Dekel, and Venkatesh Saligrama.
\newblock Adaptive neural networks for efficient inference.
\newblock In \emph{International Conference on Machine Learning (ICML)}, 2017.

\bibitem[Chen and He(2021)]{chen2021simsiam}
Xinlei Chen and Kaiming He.
\newblock Exploring simple siamese representation learning.
\newblock In \emph{IEEE Conference on Computer Vision and Pattern Recognition
  (CVPR)}, 2021.

\bibitem[Chua et~al.(2018)Chua, Calandra, McAllister, and Levine]{chua2018pets}
Kurtland Chua, Roberto Calandra, Rowan McAllister, and Sergey Levine.
\newblock Deep reinforcement learning in a handful of trials using
  probabilistic dynamics models.
\newblock In \emph{Advances in Neural Information Processing Systems
  (NeurIPS)}, 2018.

\bibitem[Dehghani et~al.(2019)Dehghani, Gouws, Vinyals, Uszkoreit, and
  Kaiser]{dehghani2019universal}
Mostafa Dehghani, Stephan Gouws, Oriol Vinyals, Jakob Uszkoreit, and Lukasz
  Kaiser.
\newblock Universal transformers.
\newblock In \emph{International Conference on Learning Representations
  (ICLR)}, 2019.

\bibitem[Elbayad et~al.(2020)Elbayad, Gu, Grave, and Auli]{elbayad2020depth}
Maha Elbayad, Jiatao Gu, Edouard Grave, and Michael Auli.
\newblock Depth-adaptive transformer.
\newblock In \emph{International Conference on Learning Representations
  (ICLR)}, 2020.

\bibitem[Fedus et~al.(2022)Fedus, Zoph, and Shazeer]{fedus2022switch}
William Fedus, Barret Zoph, and Noam Shazeer.
\newblock Switch transformers: Scaling to trillion parameter models with simple
  and efficient sparsity.
\newblock \emph{Journal of Machine Learning Research (JMLR)}, 23, 2022.

\bibitem[Figurnov et~al.(2017)Figurnov, Collins, Zhu, Zhang, Huang, Vetrov, and
  Salakhutdinov]{figurnov2017spatially}
Michael Figurnov, Maxwell~D. Collins, Yukun Zhu, Li~Zhang, Jonathan Huang,
  Dmitry Vetrov, and Ruslan Salakhutdinov.
\newblock Spatially adaptive computation time for residual networks.
\newblock In \emph{IEEE Conference on Computer Vision and Pattern Recognition
  (CVPR)}, 2017.

\bibitem[Gelada et~al.(2019)Gelada, Kumar, Buckman, Nachum, and
  Bellemare]{gelada2019deepmdp}
Carles Gelada, Saurabh Kumar, Jacob Buckman, Ofir Nachum, and Marc~G.
  Bellemare.
\newblock {DeepMDP}: Learning continuous latent space models for representation
  learning.
\newblock In \emph{International Conference on Machine Learning (ICML)}, 2019.

\bibitem[Graves(2016)]{graves2016act}
Alex Graves.
\newblock Adaptive computation time for recurrent neural networks.
\newblock \emph{arXiv preprint arXiv:1603.08983}, 2016.

\bibitem[Grill et~al.(2020)Grill, Strub, Altch{\'e}, Tallec, Richemond,
  Buchatskaya, Doersch, Pires, Guo, Azar, Piot, Kavukcuoglu, Munos, and
  Valko]{grill2020byol}
Jean-Bastien Grill, Florian Strub, Florent Altch{\'e}, Corentin Tallec,
  Pierre~H. Richemond, Elena Buchatskaya, Carl Doersch, Bernardo~Avila Pires,
  Zhaohan~Daniel Guo, Mohammad~Gheshlaghi Azar, Bilal Piot, Koray Kavukcuoglu,
  R{\'e}mi Munos, and Michal Valko.
\newblock Bootstrap your own latent: A new approach to self-supervised
  learning.
\newblock In \emph{Advances in Neural Information Processing Systems
  (NeurIPS)}, 2020.

\bibitem[Ha and Schmidhuber(2018)]{ha2018world}
David Ha and J{\"u}rgen Schmidhuber.
\newblock Recurrent world models facilitate policy evolution.
\newblock In \emph{Advances in Neural Information Processing Systems
  (NeurIPS)}, 2018.

\bibitem[Hafner et~al.(2019)Hafner, Lillicrap, Fischer, Villegas, Ha, Lee, and
  Davidson]{hafner2019planet}
Danijar Hafner, Timothy Lillicrap, Ian Fischer, Ruben Villegas, David Ha,
  Honglak Lee, and James Davidson.
\newblock Learning latent dynamics for planning from pixels.
\newblock In \emph{International Conference on Machine Learning (ICML)}, 2019.

\bibitem[Hafner et~al.(2020)Hafner, Lillicrap, Ba, and
  Norouzi]{hafner2020dream}
Danijar Hafner, Timothy Lillicrap, Jimmy Ba, and Mohammad Norouzi.
\newblock Dream to control: Learning behaviors by latent imagination.
\newblock In \emph{International Conference on Learning Representations
  (ICLR)}, 2020.

\bibitem[Hafner et~al.(2023)Hafner, Pasukonis, Ba, and
  Lillicrap]{hafner2023dreamerv3}
Danijar Hafner, Jurgis Pasukonis, Jimmy Ba, and Timothy Lillicrap.
\newblock Mastering diverse domains through world models.
\newblock \emph{arXiv preprint arXiv:2301.04104}, 2023.

\bibitem[Han et~al.(2021)Han, Huang, Song, Yang, Wang, and
  Wang]{han2021dynamic}
Yizeng Han, Gao Huang, Shiji Song, Le~Yang, Honghui Wang, and Yulin Wang.
\newblock Dynamic neural networks: A survey.
\newblock \emph{IEEE Transactions on Pattern Analysis and Machine Intelligence
  (TPAMI)}, 2021.

\bibitem[Hansen et~al.(2022)Hansen, Wang, and Su]{hansen2022tdmpc}
Nicklas Hansen, Xiaolong Wang, and Hao Su.
\newblock Temporal difference learning for model predictive control.
\newblock In \emph{International Conference on Machine Learning (ICML)}, 2022.

\bibitem[Hansen et~al.(2024)Hansen, Su, and Wang]{hansen2024tdmpc2}
Nicklas Hansen, Hao Su, and Xiaolong Wang.
\newblock {TD-MPC2}: Scalable, robust world models for continuous control.
\newblock In \emph{International Conference on Learning Representations
  (ICLR)}, 2024.

\bibitem[Henderson et~al.(2018)Henderson, Islam, Bachman, Pineau, Precup, and
  Meger]{henderson2018matters}
Peter Henderson, Riashat Islam, Philip Bachman, Joelle Pineau, Doina Precup,
  and David Meger.
\newblock Deep reinforcement learning that matters.
\newblock In \emph{AAAI Conference on Artificial Intelligence}, 2018.

\bibitem[Hinton et~al.(2015)Hinton, Vinyals, and Dean]{hinton2015distilling}
Geoffrey Hinton, Oriol Vinyals, and Jeff Dean.
\newblock Distilling the knowledge in a neural network.
\newblock \emph{arXiv preprint arXiv:1503.02531}, 2015.

\bibitem[Janner et~al.(2019)Janner, Fu, Zhang, and Levine]{janner2019mbpo}
Michael Janner, Justin Fu, Marvin Zhang, and Sergey Levine.
\newblock When to trust your model: Model-based policy optimization.
\newblock In \emph{Advances in Neural Information Processing Systems
  (NeurIPS)}, 2019.

\bibitem[Kaya et~al.(2019)Kaya, Hong, and Dumitras]{kaya2019shallowdeep}
Yigitcan Kaya, Sanghyun Hong, and Tudor Dumitras.
\newblock Shallow-deep networks: Understanding and mitigating network
  overthinking.
\newblock In \emph{International Conference on Machine Learning (ICML)}, 2019.

\bibitem[Kingma and Ba(2015)]{kingma2015adam}
Diederik~P. Kingma and Jimmy Ba.
\newblock Adam: A method for stochastic optimization.
\newblock In \emph{International Conference on Learning Representations
  (ICLR)}, 2015.

\bibitem[Lambert et~al.(2022)Lambert, Pister, and
  Calandra]{lambert2022compounding}
Nathan Lambert, Kristofer Pister, and Roberto Calandra.
\newblock Investigating compounding prediction errors in learned dynamics
  models.
\newblock \emph{arXiv preprint arXiv:2203.09637}, 2022.

\bibitem[LeCun(2022)]{lecun2022path}
Yann LeCun.
\newblock A path towards autonomous machine intelligence.
\newblock \emph{Open Review preprint (version 0.9.2)}, 2022.

\bibitem[Lee et~al.(2015)Lee, Xie, Gallagher, Zhang, and Tu]{lee2015deeply}
Chen-Yu Lee, Saining Xie, Patrick Gallagher, Zhengyou Zhang, and Zhuowen Tu.
\newblock Deeply-supervised nets.
\newblock In \emph{Artificial Intelligence and Statistics (AISTATS)}, 2015.

\bibitem[Loshchilov and Hutter(2019)]{loshchilov2019adamw}
Ilya Loshchilov and Frank Hutter.
\newblock Decoupled weight decay regularization.
\newblock In \emph{International Conference on Learning Representations
  (ICLR)}, 2019.

\bibitem[Lu et~al.(2026)Lu, Wei, Zhang, Zeng, Cao, et~al.]{lu2026loopedwm}
Hongyuan~Adam Lu, Z.~L.~Victor Wei, Qun Zhang, Jinrui Zeng, Bowen Cao, et~al.
\newblock Looped world models.
\newblock \emph{arXiv preprint arXiv:2606.18208}, 2026.

\bibitem[Raposo et~al.(2024)Raposo, Ritter, Richens, Lillicrap,
  Veli{\v{c}}kovi{\'c}, and Santoro]{raposo2024mod}
David Raposo, Sam Ritter, Blake Richens, Timothy Lillicrap, Petar
  Veli{\v{c}}kovi{\'c}, and Adam Santoro.
\newblock Mixture-of-{Depths}: Dynamically allocating compute in
  transformer-based language models.
\newblock \emph{arXiv preprint arXiv:2404.02258}, 2024.

\bibitem[Schrittwieser et~al.(2020)Schrittwieser, Antonoglou, Hubert, Simonyan,
  Sifre, Schmitt, Guez, Lockhart, Hassabis, Graepel, Lillicrap, and
  Silver]{schrittwieser2020muzero}
Julian Schrittwieser, Ioannis Antonoglou, Thomas Hubert, Karen Simonyan,
  Laurent Sifre, Simon Schmitt, Arthur Guez, Edward Lockhart, Demis Hassabis,
  Thore Graepel, Timothy Lillicrap, and David Silver.
\newblock Mastering {Atari}, {Go}, chess and shogi by planning with a learned
  model.
\newblock \emph{Nature}, 588\penalty0 (7839), 2020.

\bibitem[Schuster et~al.(2022)Schuster, Fisch, Gupta, Dehghani, Bahri, Tran,
  Tay, and Metzler]{schuster2022calm}
Tal Schuster, Adam Fisch, Jai Gupta, Mostafa Dehghani, Dara Bahri, Vinh Tran,
  Yi~Tay, and Donald Metzler.
\newblock Confident adaptive language modeling.
\newblock In \emph{Advances in Neural Information Processing Systems
  (NeurIPS)}, 2022.

\bibitem[Schwarzer et~al.(2021)Schwarzer, Anand, Goel, Hjelm, Courville, and
  Bachman]{schwarzer2021spr}
Max Schwarzer, Ankesh Anand, Rishab Goel, R.~Devon Hjelm, Aaron Courville, and
  Philip Bachman.
\newblock Data-efficient reinforcement learning with self-predictive
  representations.
\newblock In \emph{International Conference on Learning Representations
  (ICLR)}, 2021.

\bibitem[Shazeer et~al.(2017)Shazeer, Mirhoseini, Maziarz, Davis, Le, Hinton,
  and Dean]{shazeer2017moe}
Noam Shazeer, Azalia Mirhoseini, Krzysztof Maziarz, Andy Davis, Quoc Le,
  Geoffrey Hinton, and Jeff Dean.
\newblock Outrageously large neural networks: The sparsely-gated
  mixture-of-experts layer.
\newblock In \emph{International Conference on Learning Representations
  (ICLR)}, 2017.

\bibitem[Szegedy et~al.(2015)Szegedy, Liu, Jia, Sermanet, Reed, Anguelov,
  Erhan, Vanhoucke, and Rabinovich]{szegedy2015going}
Christian Szegedy, Wei Liu, Yangqing Jia, Pierre Sermanet, Scott Reed, Dragomir
  Anguelov, Dumitru Erhan, Vincent Vanhoucke, and Andrew Rabinovich.
\newblock Going deeper with convolutions.
\newblock In \emph{IEEE Conference on Computer Vision and Pattern Recognition
  (CVPR)}, 2015.

\bibitem[Talvitie(2014)]{talvitie2014hallucinated}
Erik Talvitie.
\newblock Model regularization for stable sample rollouts.
\newblock In \emph{Conference on Uncertainty in Artificial Intelligence (UAI)},
  2014.

\bibitem[Tassa et~al.(2018)Tassa, Doron, Muldal, Erez, Li, de~Las~Casas,
  Budden, Abdolmaleki, Merel, Lefrancq, Lillicrap, and
  Riedmiller]{tassa2018dmc}
Yuval Tassa, Yotam Doron, Alistair Muldal, Tom Erez, Yazhe Li, Diego
  de~Las~Casas, David Budden, Abbas Abdolmaleki, Josh Merel, Andrew Lefrancq,
  Timothy Lillicrap, and Martin Riedmiller.
\newblock {DeepMind} control suite.
\newblock \emph{arXiv preprint arXiv:1801.00690}, 2018.

\bibitem[Teerapittayanon et~al.(2016)Teerapittayanon, McDanel, and
  Kung]{teerapittayanon2016branchynet}
Surat Teerapittayanon, Bradley McDanel, and H.~T. Kung.
\newblock {BranchyNet}: Fast inference via early exiting from deep neural
  networks.
\newblock In \emph{International Conference on Pattern Recognition (ICPR)},
  2016.

\bibitem[Todorov et~al.(2012)Todorov, Erez, and Tassa]{todorov2012mujoco}
Emanuel Todorov, Tom Erez, and Yuval Tassa.
\newblock {MuJoCo}: A physics engine for model-based control.
\newblock In \emph{IEEE/RSJ International Conference on Intelligent Robots and
  Systems (IROS)}, 2012.

\bibitem[Venkatraman et~al.(2015)Venkatraman, Hebert, and
  Bagnell]{venkatraman2015dad}
Arun Venkatraman, Martial Hebert, and J.~Andrew Bagnell.
\newblock Improving multi-step prediction of learned time series models.
\newblock In \emph{AAAI Conference on Artificial Intelligence}, 2015.

\end{thebibliography}

\appendix
\section{Reproducibility}\label{app:repro}
{\raggedright Every number traces to a JSON file of results committed to the project repository, and every figure
and generated table is produced from those files by scripts in the same repository. Training uses
matched, model-independent, pre-collected data so that $K{=}1/K{=}4$ and standard/ablation twins
differ only in their loss. The routing comparison is produced by
\texttt{scripts/audit\_routing\_frontier.py} (raw per-checkpoint results in
\texttt{docs/results/routing\_frontier.json}) and summarised by
\texttt{scripts/analyze\_routing\_frontier.py}. The schedule search, trained routers, equal-compute
planner runs and supervision-matched control are produced by \texttt{audit\_schedule\_null.py},
\texttt{audit\_learned\_routers.py}, \texttt{run\_planner\_equal\_compute.py} and
\texttt{run\_ablation\_control.sh} with their \texttt{analyze\_*} counterparts. The tables in
Sections~\ref{sec:taxonomy}--\ref{sec:planner} that draw on these results are generated by
\texttt{scripts/make\_routing\_tables.py} and are not edited by hand. The code, result files and
full commit history are available from the author on request.\par}

\section{Architecture and hyperparameters}\label{app:hp}
Encoder: 2-layer MLP, latent dim 64, hidden 256. Predictor: 12 residual blocks, $d_{\text{model}}
{=}256$, MLP ratio 2, early-exit heads every 2 blocks (depths $2,4,\dots,12$), action embedding 64.
Loss: latent MSE against an EMA target (momentum $0.99$) at full depth; deep supervision of every
shallow exit; distillation weight $0.5$; VICReg variance/covariance regularization. Optimization:
AdamW \citep{kingma2015adam,loshchilov2019adamw}, lr $3\times10^{-4}$, weight decay $10^{-4}$, batch
256, 60k steps (80k for humanoid), 1000 warmup, grad clip 1. Data: 80 episodes of length 1000 from a
random policy per task, collected once per seed. Audits: $H{=}10$ ($30,50$ for the horizon study),
256 held-out windows, eval seed 456, probe-fit seed 123. Pre-registered training runs, all on H100/H200 GPUs: the
nine-task taxonomy at eight seeds ($9\times K\!\in\!\{1,4\}\times 8=144$), first-step-only ablations
(cheetah $n{=}8$, walker and quadruped $n{=}3$), the pixel replication (3 tasks $\times K \times 3 =
18$), the scaled and transformer backbone campaigns (3 anchors $\times K \times 3 = 18$ each), the
out-of-sample held-out campaign (6 tasks $\times K \times 3 = 36$), and the on-policy replication (2
tasks $\times K \times 3 = 12$)---$260$ training runs, every campaign pre-registered before its
compute. The supervision-matched control of Section~\ref{sec:catch22} adds nine runs (three arms,
three seeds), trained afterwards on an Apple-silicon GPU and not pre-registered. An earlier reduced-scale exploration and thirteen initial
ablation/expansion seeds were run on Apple-silicon and later superseded by the uniform-hardware
campaigns above. The planning, calibrated-router, horizon/stability and routing analyses are inference-time or
audit-only on frozen checkpoints, not additional training. The scaled backbone uses latent $128$,
$d_{\text{model}}\,512$, depth $24$ (exits every $4$); the transformer predictor tokenizes $[z;a]$
into $8$ tokens processed by $12$ pre-LN self-attention blocks ($4$ heads) with mean-pooled exit
heads.

\section{Why the random-data regime is the relevant one}\label{app:randomdata}
A fair objection follows from the training-distribution result of Section~\ref{sec:cautions}: if the
regime washes out once the model trains on competent-policy data, why diagnose it on
random-exploration data at all? We offer an argument, not a measurement. The decisions this
diagnostic informs---whether to add exit heads, how deep to make the predictor, whether routing can
pay---are made \emph{during model development}: architecture search, pretraining, and the early phase
of model-based RL, when no competent policy yet exists and the data is broad and exploratory. That is
precisely the regime in which the phenomenon is present, so even where the inversion fades at
convergence, the random-data taxonomy may still be the right instrument for the phase in which
depth-routing decisions are actually taken. The argument is strongest for offline and pretraining
settings; in fully online model-based RL, where the replay buffer drifts toward competence, its force
decays with the exploration it depends on, which we do not test.

\section{Pre-registration record}\label{app:prereg}
The commits named here are in the project's git history (available from the author on request) and can
be checked against the dates of the result files. The regime thresholds, Gates A/B and the metric-artifact kill criterion were committed
on 2026-07-09 (\texttt{3462e54}, \texttt{935bcb1}), before the campaign launch commit
(\texttt{387943d}). Later pre-declarations: the eight-seed expansion and confirmatory reruns
(\texttt{5efda16}, 2026-07-10), the calibrated-router protocol P2$'$ (\texttt{0995920},
2026-07-10), the out-of-sample predictor and its frozen held-out predictions (\texttt{84402b6},
2026-07-10), and the second confirmation program (\texttt{384fe52}, 2026-07-17). In
more detail: Two later additions were pre-declared
before running: the seed-count expansion of the headline cheetah comparisons (Gate B $\Delta$ and
planning return) from three seeds to eight, committed 37\,min before the first expanded checkpoint
was written; and the calibrated-router protocol P2$'$ with its disjoint calibration/evaluation split.
The out-of-sample regime predictor and its held-out predictions were frozen in a separate commit
before those tasks were trained. A second, larger confirmation program---the full nine-task
eight-seed expansion (retrained on uniform H100/H200 hardware, superseding two hardware-mixed
ablation seeds), the scaled and transformer backbone campaigns, the stronger planner, and the
on-policy replication---was pre-registered as one document before any of its compute ran, with its
confirmation criteria (e.g.\ the $\ge2/3$-dichotomy backbone criterion) fixed in advance. The full
chronological record, including the failed Gate~1 and the reduced-scale mirage that preceded it, is
in the experiment log in the same repository. The analyses added after the reviews of an earlier
version were not pre-registered: the equal-compute routing comparison, schedule search and trained
routers (Section~\ref{sec:routing}), the one-step decomposition and the supervision-matched control
(Section~\ref{sec:catch22}), and the equal-compute planner runs (Section~\ref{sec:planner}). All but
the control, which trains new models, were run on the frozen checkpoints of the pre-registered
campaigns.

\section{Regime correlates}\label{app:correlates}
\begin{figure}[H]
\centering
\includegraphics[width=0.82\linewidth]{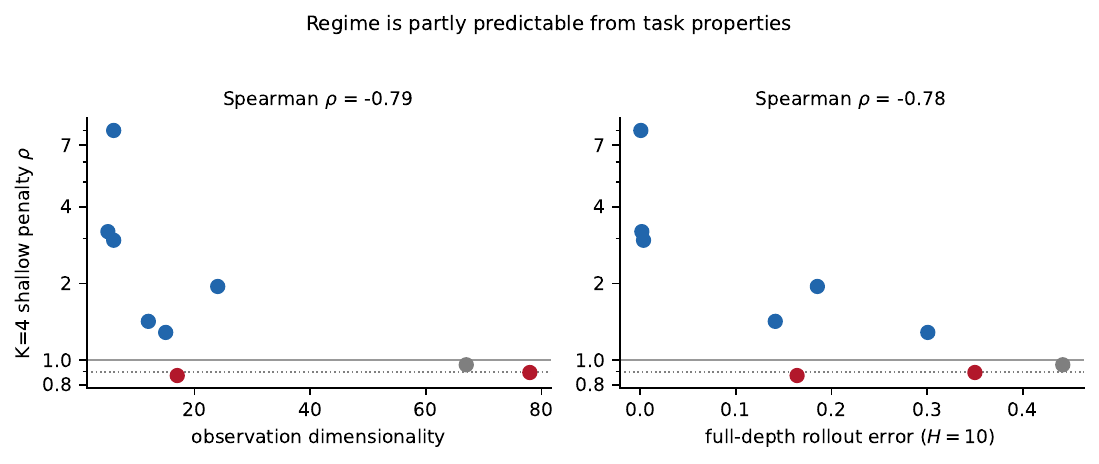}
\caption{\textbf{The regime is partly predictable.} $K{=}4$ shallow penalty vs.\ observation
dimensionality (left) and full-depth rollout error (right), colored by regime. Inversion/flat tasks
are high-dimensional and hard to model; strong intrinsic tasks are low-dimensional and accurately
modeled ($y$-axis log-scaled; $n{=}9$ tasks, eight-seed means).}
\end{figure}

\section{Planning returns}\label{app:planner}
\begin{figure}[H]
\centering
\includegraphics[width=0.54\linewidth]{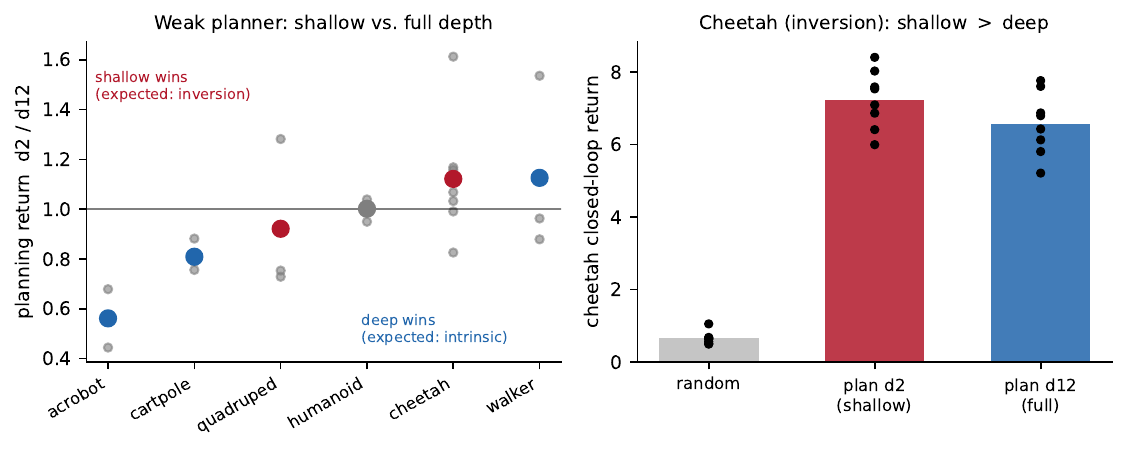}
\caption{\textbf{Weak planner: the sign of $\rhov$ and closed-loop planning (P1).} Left: ratio of shallow (d2)
to full-depth (d12) closed-loop return per task, colored by regime; $<1$ means deep planning wins
(expected for intrinsic), $>1$ shallow wins (expected for inversion). The two robust cases (cartpole,
cheetah) fall on the predicted side; the two marginal tasks (walker weak-reward, quadruped
borderline) are the misses. Right: cheetah closed-loop return---planning at depth 2 exceeds depth 12
with this planner; the stronger planner of Section~\ref{sec:planner} reverses it.}
\label{fig:planner}
\end{figure}

\begin{table}[H]
\centering
\small
\begin{tabular}{lccccc}
\toprule
 & & \multicolumn{2}{c}{original run} & \multicolumn{2}{c}{refit heads}\\
\cmidrule(lr){3-4}\cmidrule(lr){5-6}
task & seed & depth-12 return & reward $R^2$ & depth-12 return & reward $R^2$\\
\midrule
cheetah-run & 0 & 33.9 & 0.92 & 27.5 & 0.90\\
cheetah-run & 1 & 50.7 & 0.92 & 33.4 & 0.91\\
cheetah-run & 2 & 37.3 & 0.94 & 37.4 & 0.90\\
quadruped-walk & 0 & 103.7 & 0.96 & 64.5 & 0.93\\
quadruped-walk & 1 & 167.0 & 0.96 & 168.2 & 0.95\\
quadruped-walk & 2 & 172.2 & 0.94 & 169.0 & 0.92\\
\bottomrule
\end{tabular}
\caption{\textbf{Reproducing the strong-planner depth-12 runs} with refit reward heads (equal
candidates, three episodes per seed). Quadruped seed 0 is excluded from Table~\ref{tab:planner_ec}
because its refit depth-12 planner does not beat random.}
\label{tab:planner_repro}
\end{table}

\section{Metric-space dependence}\label{app:metric}
\begin{figure}[H]
\centering
\includegraphics[width=0.44\linewidth]{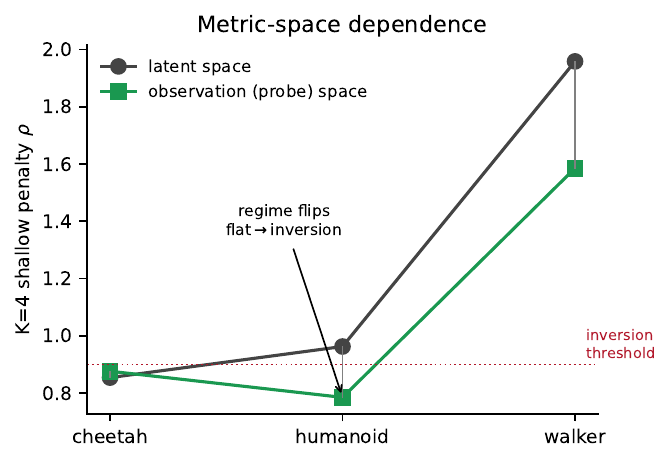}
\caption{\textbf{Metric-space dependence.} $K{=}4$ shallow penalty measured in latent vs.\
observation (probe) space; humanoid's regime flips flat$\to$inversion across the $0.90$ threshold,
while cheetah and walker are stable.}
\label{fig:metric}
\end{figure}

\section{Perturbation axes: horizon, encoder, backbone}\label{app:axes}
\begin{figure}[H]
\centering
\includegraphics[width=0.45\linewidth]{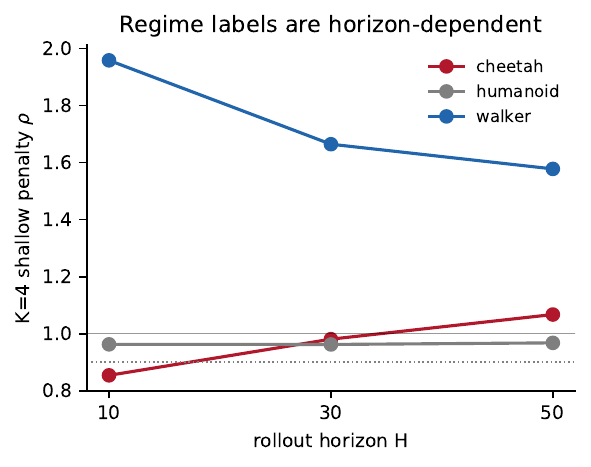}\hfill
\includegraphics[width=0.44\linewidth]{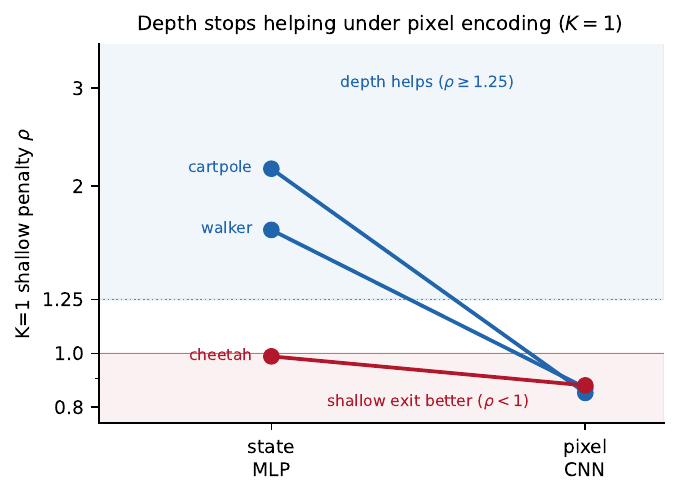}
\caption{\textbf{Left: horizon dependence.} $K{=}4$ shallow penalty vs.\ rollout horizon $H$;
cheetah's inversion recovers toward and past $1$ as errors compound, while walker and humanoid are
horizon-stable. \textbf{Right: representation dependence.} $K{=}1$ shallow penalty for the same three
tasks encoded from state (MLP) vs.\ pixels (CNN); the two intrinsic state tasks fall below 1
under pixel encoding, walker at matched one-step model quality. All three pixel tasks are flat under
the pre-registered rule, which also reads $K{=}4$.}
\end{figure}
\emph{Horizon.} Re-auditing at $H\in\{30,50\}$, cheetah's inversion recovers toward and past $1$ as
the shallow exits' small per-step errors compound, while humanoid (flat) and walker (intrinsic) are
horizon-stable. A task's regime is therefore partly a statement about the planning horizon at which
the model is used.

\emph{Encoder.} The pixel replication covers three tasks at $K\in\{1,4\}$, three seeds. The regimes do
not carry over: under the pre-registered rule all three pixel tasks are flat. At $K{=}1$, the
reading we rely on because the pixel cartpole $K{=}4$ was undertrained, the two strongly intrinsic
state tasks fall below 1 ($0.85$, $0.87$) and cheetah moves from $0.99$ to $0.88$. At $K{=}4$ the
change is milder: walker $0.97$ and cheetah $0.91$, which is \emph{less} inverted than state-based
cheetah ($0.85$ on the matched seeds). Part of this tracks model quality---the pixel cartpole is $\sim\!600\times$ harder to model, and
Section~\ref{sec:predict} finds that depth pays less on harder-to-model tasks---but not all of it,
since walker loses its depth advantage at matched one-step error. The same dynamics, encoded differently, has a different
depth-composition regime, so a task's \emph{specific} regime is a property of the (task, training,
encoder) triple rather than the task alone. (The pixel cartpole $K{=}4$ was undertrained; we
therefore read that comparison at $K{=}1$, where all six models are healthy.)

\emph{Backbone.} The backbone campaigns retrain the three anchor tasks (cheetah, walker, humanoid)
under a $2\times$-wider, $2\times$-deeper MLP predictor (latent $128$, $d_{\text{model}}\,512$, depth $24$) and a
transformer predictor (self-attention over tokens rather than a residual MLP stack), each at
$K\in\{1,4\}$, three seeds, against the pre-registered criterion that the intrinsic-vs-non-intrinsic
dichotomy match the small-backbone anchor on $\ge2/3$ tasks. Both families confirm. The one anchor
that does not carry over is walker, whose intrinsic tradeoff (at $K{=}1$) flattens under both scaling
($1.67\to1.11$) and the transformer ($\to1.08$)---the third and fourth axes, with pixels, on which
walker's \emph{magnitude} of intrinsic advantage collapses.

\section{Absolute errors and persistence}\label{app:abs}
\begin{table}[H]
\centering
\small
\setlength{\tabcolsep}{4pt}
\begin{tabular}{lcccccc}
\toprule
 & \multicolumn{2}{c}{one step} & \multicolumn{3}{c}{ten-step rollout} & \\
\cmidrule(lr){2-3}\cmidrule(lr){4-6}
task & depth 12 & persistence & depth 2 & depth 12 & persistence & skill\\
\midrule
acrobot-swingup & $1.4\!\times\!10^{-4}$ & $8.6\!\times\!10^{-3}$ & $4.5\!\times\!10^{-3}$ & $6.0\!\times\!10^{-4}$ & 0.253 & 0.998\\
cartpole-balance & $3.3\!\times\!10^{-4}$ & $6.5\!\times\!10^{-3}$ & $5.0\!\times\!10^{-3}$ & $1.7\!\times\!10^{-3}$ & 0.091 & 0.982\\
reacher-easy & $1.5\!\times\!10^{-3}$ & 0.945 & $9.5\!\times\!10^{-3}$ & $3.4\!\times\!10^{-3}$ & 1.234 & 0.997\\
walker-walk & 0.044 & 0.769 & 0.360 & 0.185 & 1.706 & 0.891\\
finger-turn\_hard & 0.089 & 1.163 & 0.199 & 0.141 & 1.377 & 0.898\\
hopper-stand & 0.175 & 0.537 & 0.386 & 0.301 & 1.336 & 0.775\\
humanoid-run & 0.156 & 1.131 & 0.423 & 0.442 & 1.843 & 0.760\\
quadruped-walk & 0.226 & 1.163 & 0.313 & 0.350 & 1.861 & 0.812\\
cheetah-run & 0.052 & 0.873 & 0.142 & 0.164 & 1.834 & 0.911\\
\bottomrule
\end{tabular}
\caption{\textbf{Absolute latent errors} (mean squared error in latent space, $K{=}4$, eight-seed
means). Persistence holds the current target-encoded latent fixed: at one step it predicts
$z_{t+1}{=}z_t$, over the rollout it predicts $z_t{=}z_0$. Skill is one minus the full-depth rollout
error over the persistence rollout error. Errors are in each task's own latent space and are not
comparable across tasks.}
\label{tab:abs}
\end{table}

\section{Routing: all tasks and further detail}\label{app:routing}
\begin{figure}[H]
\centering
\includegraphics[width=\linewidth]{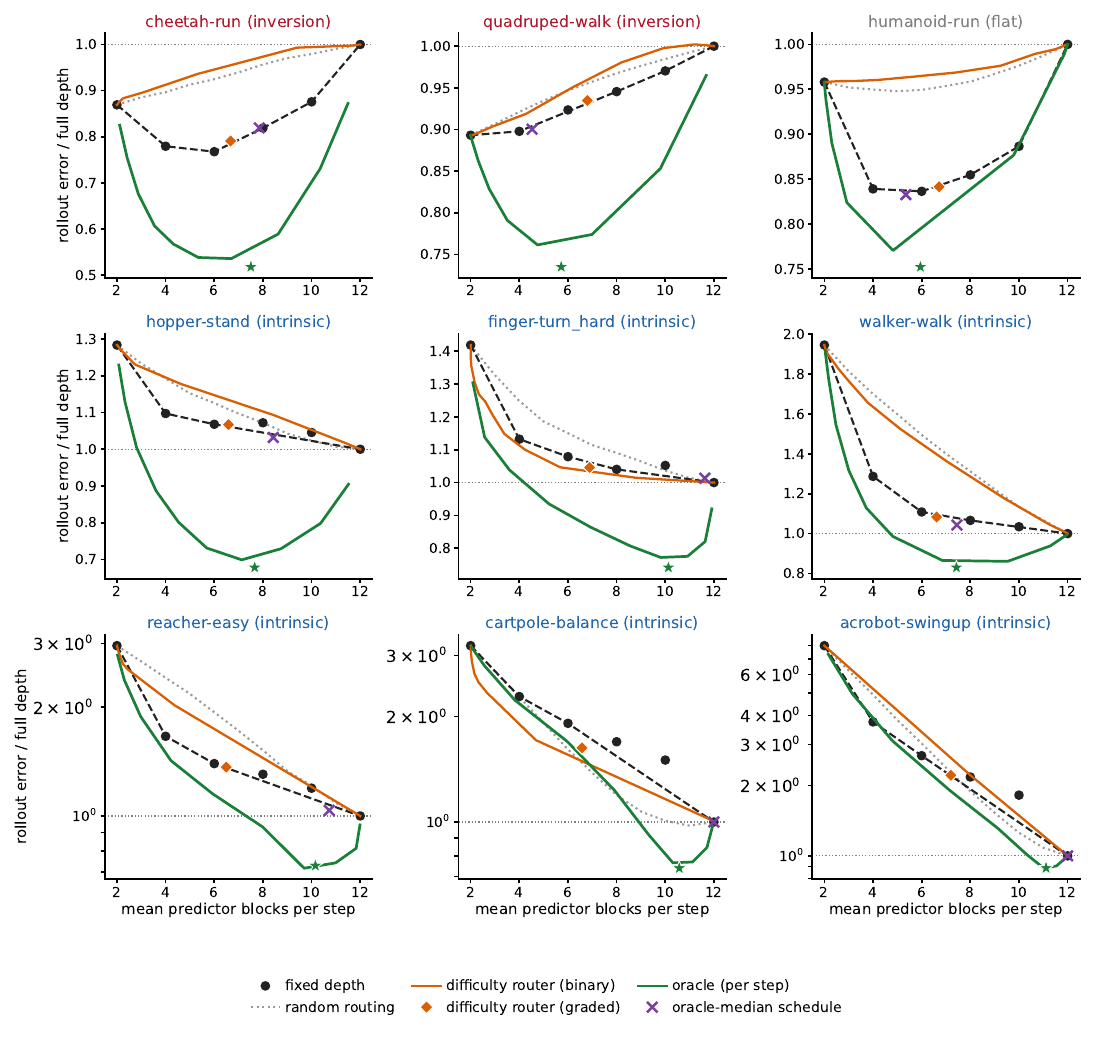}
\caption{\textbf{Rollout error against compute for all nine tasks} ($K{=}4$, eight-seed means,
relative to full depth), drawn as in Figure~\ref{fig:frontier}. Tasks with a wide error range are on a
log scale.}
\end{figure}

\begin{figure}[H]
\centering
\includegraphics[width=0.86\linewidth]{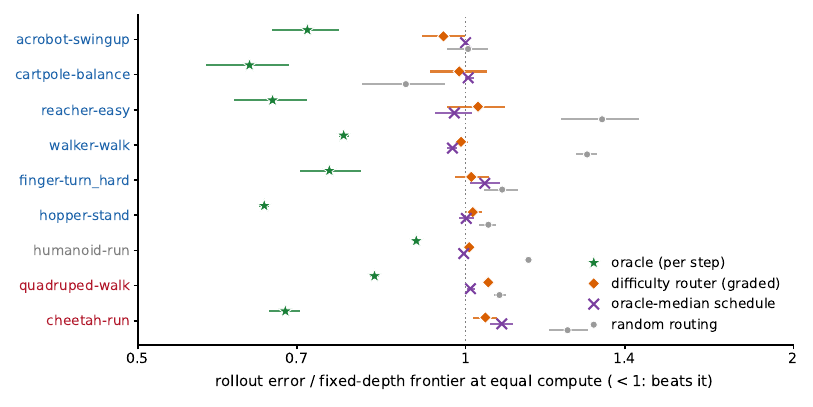}
\caption{\textbf{Routing at equal compute}, the data of Table~\ref{tab:routing} with bootstrap 95\%
CIs. Below 1, the policy beats every mixture of fixed depths at no more than its own compute.}
\end{figure}

\begin{table}[H]
\centering
\small
\setlength{\tabcolsep}{3pt}
\resizebox{\linewidth}{!}{%
\begin{tabular}{lccccccccc}
\toprule
 & \multicolumn{6}{c}{\% of steps where each exit has the lowest error} & shallower & \multicolumn{2}{c}{head Spearman with}\\
\cmidrule(lr){2-7}\cmidrule(lr){9-10}
task & 2 & 4 & 6 & 8 & 10 & 12 & beats 12 (\%) & full-depth error & depth-2 minus depth-12 error\\
\midrule
acrobot-swingup & 2 & 3 & 7 & 12 & 17 & 60 & 40 & $0.52$ $[0.46, 0.56]$ & $0.53$ $[0.50, 0.56]$\\
cartpole-balance & 5 & 5 & 7 & 11 & 16 & 55 & 45 & $0.55$ $[0.52, 0.58]$ & $0.22$ $[0.16, 0.27]$\\
reacher-easy & 1 & 3 & 8 & 16 & 27 & 45 & 55 & $0.43$ $[0.38, 0.47]$ & $0.42$ $[0.39, 0.46]$\\
walker-walk & 18 & 14 & 13 & 15 & 18 & 22 & 78 & $0.36$ $[0.34, 0.38]$ & $0.11$ $[0.10, 0.13]$\\
finger-turn\_hard & 4 & 4 & 6 & 12 & 27 & 48 & 52 & $0.67$ $[0.65, 0.69]$ & $0.37$ $[0.35, 0.40]$\\
hopper-stand & 20 & 12 & 11 & 12 & 18 & 28 & 72 & $0.40$ $[0.38, 0.41]$ & $0.06$ $[0.03, 0.08]$\\
humanoid-run & 26 & 20 & 19 & 15 & 12 & 9 & 91 & $0.12$ $[0.11, 0.13]$ & $-0.02$ $[-0.03, -0.00]$\\
quadruped-walk & 30 & 18 & 14 & 12 & 12 & 15 & 85 & $0.45$ $[0.43, 0.47]$ & $-0.07$ $[-0.08, -0.05]$\\
cheetah-run & 21 & 13 & 13 & 12 & 16 & 25 & 75 & $0.29$ $[0.27, 0.30]$ & $0.10$ $[0.08, 0.11]$\\
\bottomrule
\end{tabular}}
\caption{\textbf{Per-step preferences and what the difficulty head tracks} ($K{=}4$, eight-seed
means). Steps are taken along the full-depth rollout of each held-out window. The head correlates
with how large the full-depth error is, which it was trained to predict, much more than with whether
the deep exit beats the shallow one, which routing needs.}
\label{tab:prefs}
\end{table}

At $K{=}1$ the picture is the same: the lowest-error oracle has $0.59$--$0.87$ of the
fixed-depth frontier error, and the graded router $0.97$--$1.17$
(\texttt{docs/results/routing\_frontier\_summary.json}). An earlier single-seed check (\texttt{docs/results/e2e\_router.json}, older
checkpoints, cheetah, walker and humanoid) trained two routers: a head on oracle depth labels, and an
end-to-end router with a penalty pulling its average depth toward the oracle's. Their rollout errors
were within 3\% of random routing evaluated at the oracle's budget ($0.97$--$1.00\times$), while
skipping as many or more blocks than the oracle (59--81\% for the oracle, 63--83\% and 76--83\% for
the two routers). With one seed and budgets matched only approximately, this is suggestive only; the
eight-seed routers of Table~\ref{tab:learned} supersede it.

\section{Routing on larger, transformer, pixel and on-policy models}\label{app:variants}
One reviewer asked whether the effects survive larger models and better data. We ran the same
equal-compute analysis on every perturbation model we have: the $2\times$-wider, $2\times$-deeper MLP
predictor (24 blocks, exits every 4), the transformer predictor, the pixel-encoder models and the
models trained on competent-policy data (three seeds each; all 66 checkpoints reproduce their stored
$\rhov$ to within $5\times10^{-7}$). Table~\ref{tab:variants} gives the $K{=}4$ results.

\begin{table}[H]
\centering
\small
\setlength{\tabcolsep}{3pt}
\resizebox{\linewidth}{!}{%
\begin{tabular}{llcccccc}
\toprule
model & task & $n$ & $\rhov$ & best depth & oracle vs.\ frontier & router (graded) & router (binary)\\
\midrule
on-policy data & cartpole-balance & 3 & $1.00$ $[0.88, 1.09]$ & 2 & $0.89$ $[0.83, 0.93]$ & $1.11$ $[1.04, 1.23]$ & $1.04$ $[0.99, 1.12]$\\
on-policy data & cheetah-run & 3 & $1.06$ $[1.01, 1.10]$ & 6 & $0.78$ $[0.77, 0.80]$ & $1.00$ $[1.00, 1.02]$ & $1.02$ $[1.00, 1.03]$\\
pixels (CNN) & cartpole-balance & 3 & $1.01$ $[0.98, 1.05]$ & 12 & $0.89$ $[0.86, 0.91]$ & $1.04$ $[1.00, 1.11]$ & $0.99$ $[0.99, 1.00]$\\
pixels (CNN) & cheetah-run & 3 & $0.91$ $[0.89, 0.93]$ & 2 & $0.85$ $[0.84, 0.86]$ & $1.06$ $[1.05, 1.08]$ & $1.03$ $[1.02, 1.04]$\\
pixels (CNN) & walker-walk & 3 & $0.97$ $[0.97, 0.98]$ & 2 & $0.84$ $[0.82, 0.84]$ & $1.01$ $[1.01, 1.02]$ & $0.99$ $[0.99, 0.99]$\\
2$\times$ MLP & cheetah-run & 3 & $0.92$ $[0.88, 0.95]$ & 4 & $0.85$ $[0.84, 0.85]$ & $1.04$ $[1.01, 1.08]$ & $1.04$ $[1.03, 1.06]$\\
2$\times$ MLP & humanoid-run & 3 & $1.06$ $[1.05, 1.08]$ & 16 & $0.94$ $[0.94, 0.95]$ & $1.00$ $[1.00, 1.00]$ & $1.03$ $[1.02, 1.04]$\\
2$\times$ MLP & walker-walk & 3 & $1.06$ $[1.03, 1.10]$ & 8 & $0.88$ $[0.87, 0.90]$ & $1.01$ $[1.00, 1.02]$ & $1.05$ $[1.03, 1.07]$\\
transformer & cheetah-run & 3 & $0.88$ $[0.81, 0.96]$ & 2 & $0.81$ $[0.79, 0.83]$ & $1.09$ $[1.05, 1.16]$ & $1.05$ $[1.04, 1.07]$\\
transformer & humanoid-run & 3 & $0.94$ $[0.93, 0.94]$ & 2 & $0.87$ $[0.86, 0.87]$ & $1.06$ $[1.05, 1.06]$ & $1.02$ $[1.01, 1.02]$\\
transformer & walker-walk & 3 & $1.19$ $[1.15, 1.24]$ & 12 & $0.82$ $[0.81, 0.82]$ & $0.99$ $[0.99, 1.00]$ & $1.01$ $[1.01, 1.02]$\\
\bottomrule
\end{tabular}}
\caption{\textbf{Routing at equal compute on the perturbation models} ($K{=}4$, means over three
seeds with bootstrap 95\% CIs; ``vs.\ frontier'' as in Table~\ref{tab:routing}). The scaled model has
24 blocks, so its router runs about twice as many blocks per step.}
\label{tab:variants}
\end{table}

The fixed-depth effects weaken, as the reviewer expected: walker's strongly intrinsic ratio falls
from $1.95$ to $1.06$ under the larger MLP and to $1.19$ under the transformer, and on competent-policy
data cheetah no longer inverts. The routing results do not change. The oracle beats the
fixed-depth frontier on all eleven model--task pairs ($0.78$--$0.94$, every confidence interval
below 1). The graded router is within $1\%$ of the frontier or worse on every pair ($0.99$--$1.11$),
the binary router lands at $0.99$--$1.05$, and the difficulty head's correlation with
the depth-2 minus depth-12 gap is $-0.19$ to $0.03$.

\section{All nine depth-curve profiles}\label{app:grid}
\begin{figure}[H]
\centering
\includegraphics[width=\linewidth]{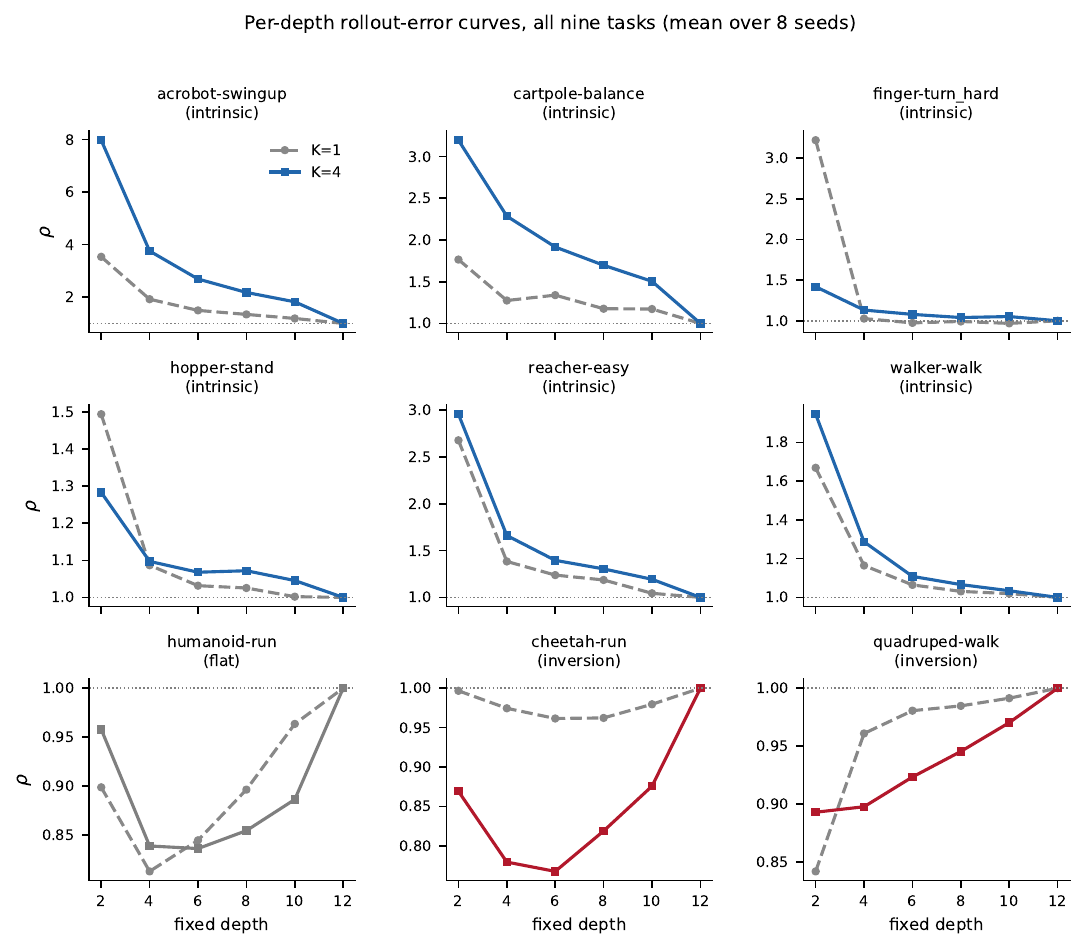}
\caption{\textbf{Per-depth rollout-error curves for all nine tasks} (mean over seeds; $K{=}1$ dashed
grey, $K{=}4$ solid, colored by regime). Intrinsic tasks fall with depth, nearly monotonically; inversion
tasks (cheetah, quadruped) dip below 1 at intermediate depths; humanoid (flat by the shallowest-exit
ratio) nonetheless shows an inversion-shaped dip to $\approx0.84$ at depths 4--6, illustrating why its
regime is borderline and metric-dependent (Section~\ref{sec:cautions}).}
\end{figure}

\section{Contractivity (null mechanism)}\label{app:contractivity}
\begin{figure}[H]
\centering
\includegraphics[width=\linewidth]{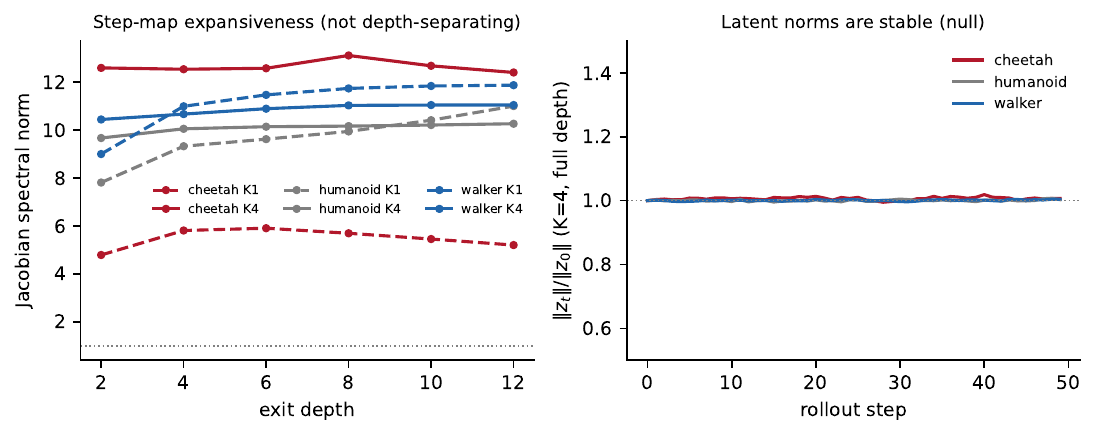}
\caption{\textbf{The inversion is not a contractivity effect.} Left: mean Jacobian spectral norm of
the one-step map vs.\ exit depth (all $>1$, i.e.\ locally expansive; $K{=}4$ raises and flattens
cheetah's, but the norms do not separate shallow from deep in a regime-predictive way). Right:
$\|z_t\|/\|z_0\|$ over a 50-step full-depth self-rollout is flat at $\approx1.0$ for all tasks---no
blow-up or collapse. The inversion is a training effect (Section~\ref{sec:catch22}), not a
fixed-point property of the maps.}
\end{figure}

\end{document}